\documentclass[11pt, a4paper, logo, copyright, nonumbering]{seed}

\usepackage{natbib}

\usepackage{CJKutf8}

\usepackage{xargs}  

\usepackage{todonotes}  

\usepackage{multirow}

\usepackage{cleveref}

\usepackage{amsmath}
\usepackage{dsfont}

\usepackage{subcaption}

\usepackage{svg}

\usepackage{makecell}
\usepackage{tabularx}
\usepackage{array}
\usepackage{wrapfig}
\usepackage{subcaption}
\usepackage{multirow}
\usepackage{multicol}
\usepackage{setspace}

\newcommand{\benchmark}{{FullStack Bench}}

\renewcommand{\today}{}
\newcommandx{\info}[2][1=]{\todo[linecolor=red,backgroundcolor=red!25,bordercolor=red,#1]{#2}}

\title{\centering FullStack Bench: Evaluating LLMs as Full Stack Coders}

\author{\textbf{Bytedance, Seed Foundation Code Team}}

\begin{abstract}

\end{abstract}

\begin{abstract}
As the capabilities of code large language models (LLMs) continue to expand, their applications across diverse code intelligence domains are rapidly increasing. However, most existing datasets only evaluate limited application domains. To address this gap, we have developed a comprehensive code evaluation dataset \benchmark{}\footnote{\url{https://huggingface.co/datasets/ByteDance/FullStackBench}}\textsuperscript{,}\footnote{\url{https://github.com/bytedance/FullStackBench}} focusing on full-stack programming,
which encompasses a wide range of application domains (e.g., basic programming, data analysis, software engineering, mathematics, and machine learning).
Besides, to assess multilingual programming capabilities, 
in \benchmark{},  we design real-world instructions and corresponding unit test cases from 16 widely-used programming languages to reflect real-world usage scenarios rather than simple translations. 
Moreover, we also release an effective code sandbox execution tool (i.e., SandboxFusion\footnote{\url{https://github.com/bytedance/SandboxFusion}}) supporting various programming languages and packages to evaluate the performance of our \benchmark{} efficiently.
Comprehensive experimental results on our \benchmark{} demonstrate the necessity and effectiveness of our \benchmark{} and SandboxFusion.
\end{abstract}

\begin{document}

\maketitle

\let\oldthefootnote\thefootnote
\renewcommand*{\thefootnote}{\fnsymbol{footnote}}
\footnotetext[0]{Author contributions \hyperref[sec:contributions]{listed at end of paper}.}
\let\thefootnote\oldthefootnote
\footnotetext[1]{\url{https://huggingface.co/datasets/ByteDance/FullStackBench}}
\footnotetext[2]{\url{https://github.com/bytedance/FullStackBench}}
\footnotetext[3]{\url{https://github.com/bytedance/SandboxFusion}}

\begin{figure}[!ht]
    \centering
    \includegraphics[width=\linewidth]{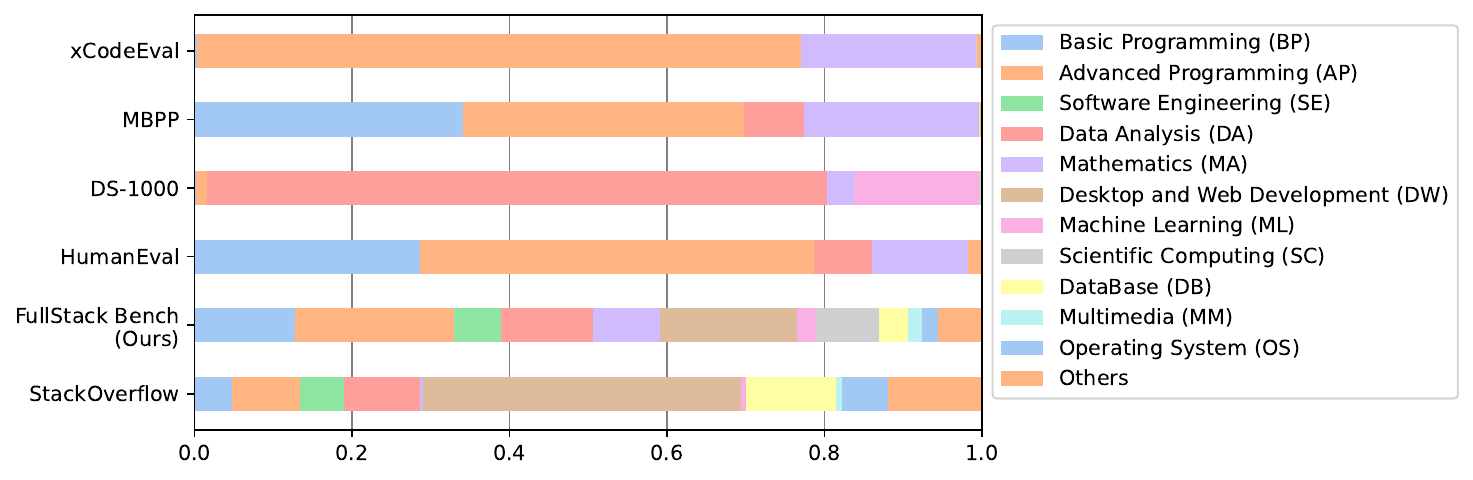}
    \caption{Application domain distributions of different code evaluation datasets.}
    \label{fig:compare}
\end{figure} 
\newpage

\tableofcontents

\newpage

\section{Introduction}

The code large language models (LLMs) have achieved significant improvements in code intelligence~\citep{codellama,codegeex,deepseek_coder,qwen25coder,Huang2024OpenCoderTO},
which are pre-trained on extensive datasets comprising billions of code-related tokens.
Recently, to discover the limitations of existing code LLMs and facilitate further development of code intelligence,
many code evaluation benchmark datasets (e.g., HumanEval~\citep{chen2021evaluatinglargelanguagemodels}, MBPP~\citep{mbpp}, DS-1000~\citep{Lai2022DS1000}, xCodeEval~\citep{khan2023xcodeeval}) have been proposed as shown in Figure~\ref{fig:compare}.

However, as shown in Figure~\ref{fig:compare}, we observe that \textbf{the existing benchmarks cover limited application domain types, which cannot access the code-related abilities of the real-world code development scenarios}.
Specifically, 
in Figure~\ref{fig:compare},
we sample 500k questions from the widely-used software development community (i.e., ``StackOverflow'') and tag the application domain label for these questions based on LLMs\footnote{Prompt: \textit{ You are an expert in the field of computer programming, proficient in various programming knowledge. Below, I will provide a set of user questions and AI assistant answers. You need to output application domain tags for this Q\&A pair. Here are some tags for reference: Mathematics, Data Analysis, Database, Desktop and Web Development. }}.
Then, based on the labels on ``StackOverflow'', we summarize 11 main-stream application domains (e.g., Basic Programming, Software Engineering, Data Analysis),
which cover about 88.1\% problems in ``StackOverflow''.
Meanwhile,
using these domain labels,
we also tag four popular code evaluation datasets (i.e., HumanEval, MBPP, DS-1000, xCodeEval), and observe that these benchmarks usually focus on {very limited domains}. For example, a large portion of DS-1000 (>95\%) is related to data analysis and machine learning tasks, and even the so-called multi-task benchmark xCodeEval (with code understanding, generation, translation and retrieval tasks) mainly focuses on advanced programming and mathematics domains.

To address the abovementioned limitation, 
we propose the \textbf{FullStack Bench}, an evaluation set spanning multiple computer science domains and programming languages, which aims to assess large models' capabilities across various real-world code development scenarios.
As shown in Figure~\ref{fig:compare},
when compared to existing benchmarks, our FullStack Bench covers more application domains,
which demonstrates the diversity and necessity of our \benchmark{}.
Besides, based on the analysis of StackOverflow, we observe that our FullStack Bench can simulate StackOverflow well for real-world programming scenes,
where ratios of the selected 11 application domains (excluding ``Others'') for our FullStack Bench and StackOverflow are 94.3\% and 88.1\%, respectively.
Moreover,
automating the evaluation on \benchmark{}  is challenging due to the various data formats and dependencies for different application domains and programming languages.
Recently,
{some sandbox execution environments (i.e.,  DifySandbox~\citep{difysandbox}, MultiPL-E~\citep{cassano2023multiple}, MPLSandbox~\citep{dou2024multiprogramminglanguagesandboxllms}) have been proposed.
However,
\textbf{there are significant limitations (e.g., supporting limited packages and programming languages) in these sandboxes}, 
which cannot evaluate our \benchmark{} well.
For example, the front-end browsers and deep-learning packages (e.g., PyTorch~\citep{Paszke2019PyTorchAI}, Tensorflow~\citep{tensorflow2015}) are not supported in these sandboxes.
Besides, 
our \benchmark{} has 16 programming languages (i.e., Bash,	C++, C\#, D, Go, HTML, Java, Javascript, PHP, Python, R, Ruby, Rust, Scala, SQL, Typescript), and many sandboxes do not fully support these languages.
Therefore,
we also introduce a new execution environment (i.e., \textbf{SandboxFusion}) to support the evaluation on our \benchmark{},
{and the main features of SandboxFusion are as follows:
(1) \textbf{Supporting various languages}: our SandboxFusion supports 23 commonly-used programming languages, which satisfies different real-world usage scenes (e.g., front-end development, backend development, ML training)
(2) \textbf{Easy-to-deploy}: we only need a single server to deploy our SandboxFusion with high throughput for large model evaluation scenarios.
(3) \textbf{Unified multi-dataset execution environment}: Apart from our \benchmark{}, we additionally support 10+ widely-used code evaluation  benchmarks.
}


Overall, the contributions are summarized as follows:
\begin{itemize}
    \item To access the code-related abilities of the real-world code development scenarios, we propose the \benchmark{} dataset with 3374 problems for 16 programming languages,
    which covers more mainstream application domains when compared to existing code evaluation benchmarks. 
    \item Meanwhile, to evaluate our \benchmark{} efficiently, we also release an effective code sandbox execution tool (i.e., SandboxFusion) to evaluate different programming tasks from different languages.
    
\item  Comprehensive experimental results and detailed analysis demonstrate the necessity and effectiveness of our \benchmark{} and SandboxFusion. Notably, We compared the performance scores on the test set of \benchmark{} with the performances of models on HumanEval in Figure~\ref{fig:performance}.
Most models are located in the upper triangular area of the graph,
with many models scoring high on Humaneval but exhibiting relatively lower performance on \benchmark{}.
 \end{itemize}

\begin{figure}[t]
    \centering
    \includegraphics[width=0.8\linewidth]{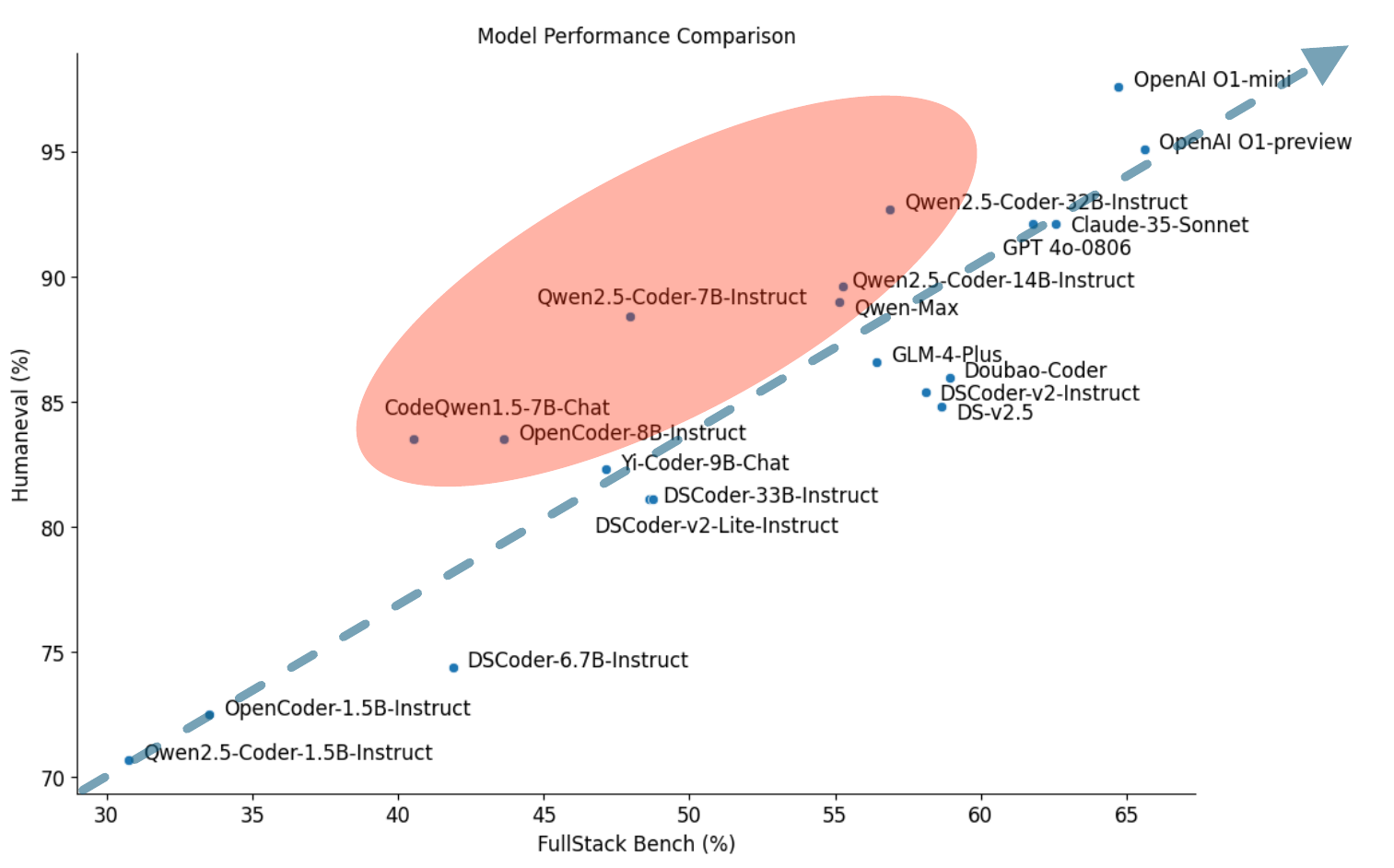}
    \caption{Performance plot of tested LLMs on  HumanEval and \benchmark{}.}
    \label{fig:performance}
\end{figure} 
\section{FullStack Bench}





\subsection{Data Overview}

As illustrated in Table~\ref{tab:detail_data}, the \benchmark{} consists of 3374 problems, where
each problem in \benchmark{} includes \textit{question, unit test cases, reference solution, and labels}. Besides, we also calculate the token lengths of the question and correct code using the LLaMA3 tokenizer~\citep{dubey2024llama3},
where {the average question length is 210.2 tokens.}
 To ensure judgment accuracy,
the overall number of unit tests for the dataset is 15168, 
where the average number of unit tests is 4.5.
We strive to cover all error types in each language. Due to the inherent differences among languages, we ensure a balanced distribution of difficulty levels, leading to variations in the distribution of error types across languages.

\subsection{Data Construction and Quality Control}
\label{sec:data_quality}
To curate the multilingual full stack code evaluation benchmark \benchmark{}, we employ a comprehensive and systematic human annotation process for producing code samples of different application domains,
where meticulously pre-defined guidelines are provided to guarantee accuracy and consistency.


\begin{figure}[t]
    \centering
    \includegraphics[width=\linewidth]{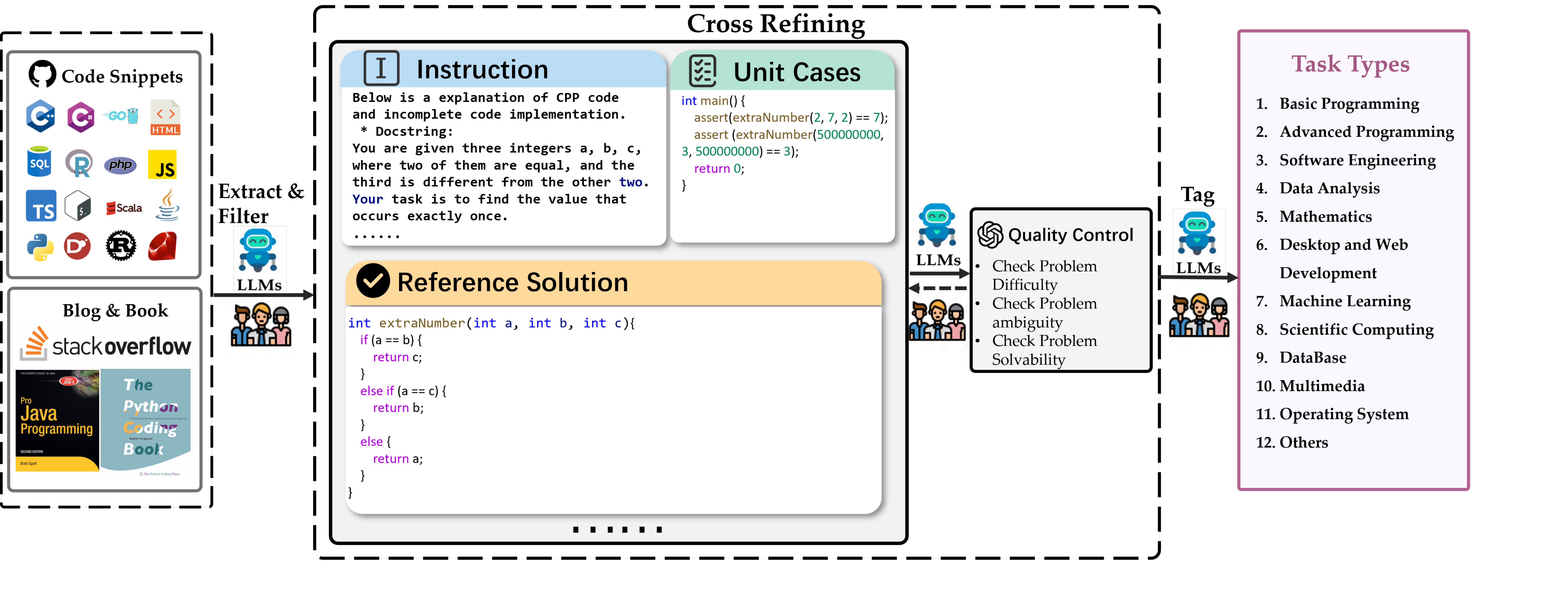}
    \caption{Overview of data collection process of \benchmark{}.}
    \label{fig:teaser-1}
\end{figure} 

\begin{wraptable}[19]{r}{0.5\textwidth}
\centering
\resizebox{0.495\textwidth}{!}{

\begin{tabular}{lr}
\hline
\toprule
\textbf{Statistics}            & \textbf{Number}       \\
\midrule
\textbf{\#Problems}                  &        $3374$          \\
\midrule
\textbf{Difficulty Level } &                          \\
- Easy/Medium/Hard               & $1,466$/$1,184$/$724$ \\
\midrule
\textbf{Length}                                         \\
Question \\
~~~~- \textit{maximum length}   & $1931$ tokens          \\
~~~~- \textit{minimum length}   & $35$ tokens            \\
~~~~- \textit{avg length}       & $210.2$ tokens         \\
Reference Solution  \\
~~~~- \textit{maximum length}   & $2720$ tokens          \\
~~~~- \textit{minimum length}   & $4$ tokens            \\
~~~~- \textit{avg length}       & $153.0$ tokens         \\
\bottomrule
\hline
\end{tabular}
}
\caption{Dataset statistics of \benchmark{}. }

\label{tab:detail_data}
\end{wraptable}

Specifically,
as shown in Figure \ref{fig:teaser-1}, we illustrate the overall dataset construction process.
Specifically, we first collect code snippets from Github, code-related documents (e.g., blog and Book) and XLCoST~\citep{zhu2022xlcost} .
Then, 
we use LLM and human verification to generate the instruction, unit cases and corresponding reference solution.
Besides,  we also employ programming experts actively in each field to create domain-specific questions for LLMs. These questions do not involve proprietary information, but are designed to assess essential skills in the respective application domains, similar to interview questions. For example, we engaged our internal data engineering team
 to develop a series of data analysis questions, including data filtering, data mining, and data visualization.
After obtaining the initial dataset, to improve the annotation quality, the annotators evaluate the annotated code based on three criteria: problem difficulty, ambiguity and solveability. Furthermore, after completing their annotations, each annotator exchanges data with another annotator for cross-refining, aiming to minimize subjective bias and errors. Any discrepancies between annotators are resolved through consensus or with input from senior annotators. 



Additionally, to improve the difficulty of our \benchmark{},
we follow the LIME~\citep{zhu2024lime} and implement a voting method using six selected models {(i.e., DeepSeek-Coder-6.7B~\citep{deepseek_coder}, DeepSeek-Coder-33B~\citep{deepseek_coder}, Qwen2.5-Coder-7B~\citep{qwen25coder}, LLaMA3.1-70B~\citep{dubey2024llama3}, Claude-3.5-Sonnet~\footnote{\url{https://www.anthropic.com/news/claude-3-5-sonnet}}, GPT-4o~\citep{achiam2023gpt4})} to filter out samples that can be correctly answered by all these LLMs.
Specifically,
for each question,
if only one model obtains the correct answer, this question is classified as a hard sample,
and if five or six models obtain the correct answer, this question is classified as an easy sample.
Apart from the easy and hard samples,
the difficulty of the remained samples is medium.


Moreover, to simulate real-world usage of full-stack developments, 
we summarize the common application domains by analyzing the distributions of ``Stackoverflow.com''.
As shown in Figure~\ref{fig:teaser},
we sample 500k questions from ``Stackoverflow.com'' and then prompt the LLMs to label the application domain type for each question.
After that, we preserve the top 11 application domains, which dominate 88.1\% of the questions. Meanwhile, we name other application domain types as ``Others''.
In this way, we also prompt GPT to label the domain types of our annotated questions and generate our final FullStack Bench,
where the domain types are as follows:

\begin{itemize}
\item \textbf{Basic Programming (BP)}: Basic programming involves fundamental concepts and skills to write simple computer programs. This typically includes understanding data types, variables, control structures, functions, and basic input/output operations.

\item \textbf{Advanced Programming (AP)}: Advanced programming involves developing complex software solutions and focuses on creating efficient, scalable, and robust applications while implementing sophisticated algorithms, data structures, and design patterns.

\item \textbf{Software Engineering (SE)}: Software engineering covers the design, development, testing, and maintenance of software systems, and includes tasks of requirements analysis, software architecture design, coding, quality assurance, and project management.

\item \textbf{Data Analysis (DP)}: Data analysis is the cleaning, processing, and analysis of collected data to discover meaningful patterns and relationships to make data-driven decisions.

\item \textbf{Mathematics (MA)}: Mathematical problems involve solving various problems through mathematical methods and theories, covering multiple fields such as algebra, geometry, calculus, number theory, probability, and statistics

\item \textbf{Desktop and Web Development (DW)}: Desktop development encompasses a wide range of programming languages, frameworks, and tools to design, build, and maintain user-friendly interfaces and robust backend systems.

\item \textbf{Machine Learning (ML)}: Machine learning algorithms are developed to learn from data for tasks such as classification, prediction, and pattern recognition.

\item \textbf{Scientific Computing (SC)}: Scientific computing solves complex scientific and engineering problems, which encompasses the tasks of numerical analysis, and high-performance computing to simulate, model, and analyze phenomena across various scientific disciplines.

\item \textbf{DataBase (DB)}: Database includes tasks such as insertion, querying, updating, and deletion, and these tasks are typically performed using query languages such as SQL to ensure efficient storage and retrieval of data.

\item \textbf{Multimedia (MM)}: Multimedia involves processing and manipulating various forms of content, including text, images, audio, and video.

\item \textbf{Operating System (OS)}: Operating system includes tasks such as memory management, process scheduling, file system management, and device control,
which aims to manage computer hardware and software resources.

\item \textbf{Others}: Apart from the above 11 mainstream application domains, other domains are categorized as ``Others''.
\end{itemize}


\subsection{Bilingual Benchmark Construction}
The collected questions are in Chinese or English. For Chinese or English problems, we translate these problems into English or Chinese, which results in both Chinese and English versions. Finally, in \benchmark{}, the numbers of Chinese and English problems are both $3374/2=1687$.

\subsection{Evaluation Metrics}

Following HumanEval and MBPP, we directly use the Pass@1 as the default evaluation metric for our proposed \benchmark{}.


\section{SandboxFusion}

Execution-based datasets are crucial for discriminating code generation tasks ~\citep{hendrycks2021measuring}.  Automating the evaluation of these datasets requires extracting complete code from the model's responses, and executing it in a compatible environment. This is a complex task due to the varying data formats and dependencies. 
To facilitate the evaluation of \benchmark{}, we also propose the SandboxFusion execution environment. SandboxFusion is a unified architecture that is compatible with many datasets as well as Fullstack Bench. This makes the sandbox widely applicable for data processing, model evaluation, reinforcement learning, etc.


\begin{figure}[t]
    \centering
    \includegraphics[width=\linewidth]{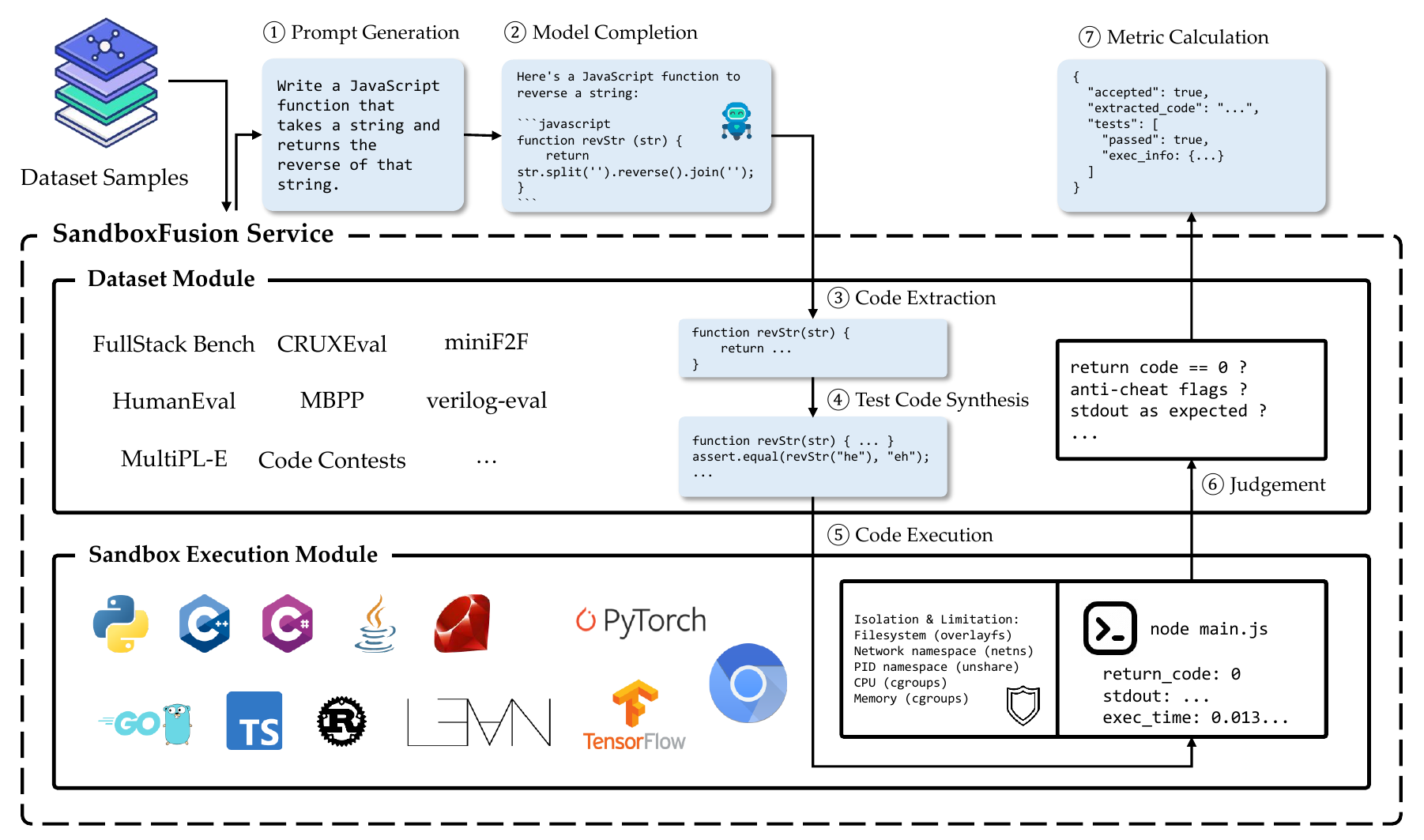}
    \caption{SandboxFusion Architecture.}
    \label{fig:teaser}
\end{figure}

As shown in Figure~\ref{fig:teaser}, the overall evaluation process of SandboxFusion usually involves the following steps:

\begin{itemize}
\item \textbf{Prompt Generation}: The system generates diverse prompts based on the original problem specifications and evaluation paradigms (e.g., few-shot, zero-shot), enabling systematic assessment of model capabilities.
\item \textbf{Model Completion}: Users need to perform model completion using the generated prompts independently, as our framework does not provide built-in inference capabilities. While many efficient inference engines exist (e.g., vLLM, text-generation-inference), we focus on prompt generation and evaluation.
\item \textbf{Code Extraction}: The system extracts executable code segments from model outputs, primarily focusing on code contained within markdown blocks.
\item \textbf{Test Code Synthesis}: The framework combines the extracted code with predefined test cases to create executable test programs. This process handles various language-specific requirements, such as distributing classes across files in Java or adapting main functions for unit testing.
\item \textbf{Code Execution}: The system executes the synthesized code with all dependent files and captures program output.
\item \textbf{Judgement}: The framework assesses solution correctness based on execution results, typically through standard unit testing frameworks where zero return values indicate successful execution.
\item \textbf{Metric Calculation}: The evaluation primarily focuses on pass rates across different problem instances.
\end{itemize}

SandboxFusion mainly contains two modules: the \textbf{Dataset Module} and the \textbf{Sandbox Execution Module}. The dataset module is responsible for implementing various datasets and abstracting out common components for reuse. The sandbox execution module focuses on executing code in different languages, controlling resource usage, and ensuring execution safety. Please See Appendix~\ref{app:sandbox} for more details of SandboxFusion and more comparisons with other sandboxes in Appendix~\ref{app:compare}.





\section{Experiments}

\begin{table}[t]
\resizebox{0.98\textwidth}{!}{
\begin{tabular}{l|ccccccccccccc}
\toprule[1.5pt]
\multicolumn{1}{c}{Model} & BP & AP & SE & DP & MA & DW & ML & SC & DB & MM & OS & Others & Overall \\
\midrule
\multicolumn{14}{c}{1B+ Instruction Tuned Coder}\\
\midrule
OpenCoder-1.5B-Instruct & 26.05 & 40.03 & 31.50 & 42.64 & 25.17 & 39.12 & 23.75 & 13.97 & 30.16 & 26.67 & 44.12 & 38.30 & \textbf{33.52} \\
Qwen2.5-Coder-1.5B-Instruct & 18.37 & 34.75 & 29.00 & 33.50 & 28.32 & 41.33 & 17.50 & 15.81 & 40.48 & 23.33 & 47.06 & 28.19 & 30.74 \\
DeepSeek-Coder-1.3B-Instruct    & 
16.74 & 29.91 & 32.50 & 37.06 & 22.73 & 35.54 & 18.75 & 9.19 & 27.78 & 25.00 & 36.76 & 30.32 & 27.65 \\
\midrule
\multicolumn{14}{c}{6B+ Instruction Tuned Coder}\\
\midrule
Qwen2.5-Coder-7B-Instruct       &
38.60 & 53.23 & 39.00 & 63.20 & 49.65 & 44.56 & 37.50 & 33.46 & 46.83 & 55.00 & 63.24 & 54.26 & \textbf{47.95} \\
Yi-Coder-9B-Chat & 
39.07 & 46.04 & 39.50 & 64.97 & 46.50 & 49.66 & 42.50 & 34.93 & 48.41 & 41.67 & 58.82 & 49.47 & 47.13 \\
OpenCoder-8B-Instruct           & 
39.53 & 49.12 & 38.00 & 55.58 & 36.01 & 45.92 & 27.50 & 26.47 & 47.62 & 46.67 & 45.59 & 45.74 & 43.63          \\
DeepSeek-Coder-7B-Instruct-v1.5 & 
38.37 & 45.16 & 36.00 & 57.36 & 35.66 & 47.96 & 30.00 & 30.88 & 46.03 & 53.33 & 45.59 & 44.15 & 43.48          \\
DeepSeek-Coder-6.7B-Instruct    & 
34.19 & 43.40 & 38.50 & 58.12 & 38.11 & 43.88 & 33.75 & 23.90 & 46.03 & 38.33 & 60.29 & 44.15 & 41.88          \\
CodeQwen1.5-7B-Chat             & 
36.74 & 44.87 & 46.00 & 51.78 & 29.72 & 40.82 & 26.25 & 24.26 & 42.06 & 41.67 & 48.53 & 44.68 & 40.52          \\
CodeLlama-7B-Instruct           & 
21.40 & 21.70 & 30.50 & 34.26 & 20.28 & 40.48 & 8.75 & 11.76 & 34.92 & 15.00 & 50.00 & 29.26 & 27.06          \\
\midrule
\multicolumn{14}{c}{13B+ Instruction Tuned Coder}\\
\midrule
Qwen2.5-Coder-14B-Instruct & 
53.26 & 58.50 & 41.00 & 69.54 & 69.23 & 46.26 & 51.25 & 43.01 & 49.21 & 60.00 & 69.12 & 57.45 & \textbf{55.28} \\
DeepSeekCoder-v2-Lite-Instruct  
& 45.81 & 57.18 & 38.50 & 56.85 & 52.80 & 44.56 & 42.50 & 33.82 & 52.38 & 33.33 & 50.00 & 51.60 & 48.73          \\
StarCoder2-15B-Instruct-v0.1    & 
38.37 & 42.23 & 29.00 & 59.90 & 37.06 & 40.99 & 42.50 & 28.68 & 54.76 & 33.33 & 42.65 & 45.74 & 41.79 \\
CodeLlama-13B-Instruct          & 
24.88 & 21.41 & 31.00 & 31.47 & 18.18 & 41.67 & 16.25 & 13.24 & 35.71 & 15.00 & 45.59 & 32.45 & 27.59          \\
\midrule
\multicolumn{14}{c}{20B+ Instruction Tuned Coder}\\
\midrule
DeepSeekCoder-v2-Instruct       
& 52.79 & 63.64 & 43.00 & 71.57 & 75.87 & 47.45 & 46.25 & 52.94 & 53.97 & 51.67 & 63.24 & 59.57 & \textbf{58.09} \\
Qwen2.5-Coder-32B-Instruct &
51.86 & 60.85 & 43.00 & 73.10 & 69.93 & 47.11 & 55.00 & 44.85 & 56.35 & 61.67 & 61.76 & 60.64 & 56.88 \\
DeepSeekCoder-33B-Instruct      
& 38.37 & 50.59 & 35.50 & 65.99 & 50.00 & 49.49 & 43.75 & 39.71 & 49.21 & 53.33 & 54.41 & 48.40 & 48.61          \\
CodeLlama-34B-Instruct          
& 23.72 & 22.73 & 26.50 & 37.56 & 18.18 & 43.71 & 17.50 & 17.65 & 38.10 & 26.67 & 51.47 & 30.85 & 29.22          \\
\midrule
\multicolumn{14}{c}{70B+ Instruction Tuned General Language Model}\\
\midrule
Qwen2.5-72B-Instruct            
& 52.56 & 61.44 & 43.00 & 66.50 & 76.57 & 48.47 & 55.00 & 51.10 & 52.38 & 51.67 & 55.88 & 55.32 & \textbf{56.88} \\
Llama3.1-70B-Instruct           
& 46.51 & 54.69 & 34.50 & 65.48 & 64.69 & 45.24 & 51.25 & 38.60 & 56.35 & 46.67 & 57.35 & 53.72 & 51.45         \\
\midrule
\multicolumn{14}{c}{Close-Sourced API Model}\\
\midrule
OpenAI o1-preview               
& 71.63 & 71.99 & 49.50 & 72.59 & 80.77 & 51.53 & 50.00 & 63.97 & 57.14 & 60.00 & 67.65 & 68.09 & \textbf{65.62} \\
OpenAI o1-mini                  
& 70.23 & 75.66 & 41.50 & 71.07 & 81.47 & 48.47 & 56.25 & 59.19 & 54.76 & 60.00 & 69.12 & 67.55 & 64.73          \\
Claude-35-Sonnet                
& 61.63 & 65.40 & 53.00 & 71.83 & 77.27 & 48.81 & 53.75 & 63.24 & 58.73 & 68.33 & 64.71 & 68.62 & 62.57          \\
GPT 4o-0806                     
& 57.21 & 67.60 & 46.00 & 74.37 & 76.92 & 48.47 & 63.75 & 55.88 & 60.32 & 63.33 & 70.59 & 64.89 & 61.77         \\
Doubao-Coder-Preview
& 56.98 & 64.66 & 43.00 & 71.07 & 74.48 & 49.15 & 45.00 & 59.19 & 50.00 & 48.33 & 60.29 & 55.32 & 58.92 \\
DeepSeek-v2.5                   
& 51.86 & 63.78 & 43.00 & 69.54 & 75.17 & 49.66 & 47.50 & 57.35 & 53.17 & 60.00 & 60.29 & 61.70 & 58.65          \\
GLM-4-Plus                      
& 49.77 & 59.97 & 43.00 & 71.32 & 72.73 & 46.43 & 56.25 & 50.00 & 57.14 & 61.67 & 63.24 & 52.66 & 56.40          \\
Qwen-Max                        
& 47.21 & 61.14 & 42.00 & 63.20 & 72.73 & 47.11 & 53.75 & 55.15 & 57.94 & 41.67 & 54.41 & 50.53 & 55.16   \\
\bottomrule
\end{tabular}
}
\caption{Model performance across domains.}
\label{tab:category}
\end{table}

\subsection{Experimental Setup}
\paragraph{FullStack AI Coders.} We select 27 popular (code) language models as full-stack AI coders and test them with \benchmark. For open-sourced models, we select AI coders from well-known and uprising code LLM series, including CodeQwen1.5~\citep{bai2023qwen}, Qwen2.5-Coder~\citep{qwen25coder}, DeepSeek-Coder~\citep{guo2024deepseek}, Deep-Seek-Coder-v2~\citep{zhu2024deepseek}, CodeLlama~\citep{codellama}, Yi-Coder~\citep{young2024yi}, StarCoder2~\citep{lozhkov2024starcoder2}, and OpenCoder~\citep{huang2024opencoder}. Further, we involve two open-sourced general LLMs, Qwen2.5 \footnote{\url{https://qwenlm.github.io/blog/qwen2.5/}} and Llama3.1~\citep{dubey2024llama3}, into the comparison. As the majority of problems in \benchmark ~are complex natural language instructions, we adopt the instruction-tuned version of those AI coders rather than their base models. According to the model size, we categorize the AI coders into five groups: 1B+, 6B+, 13B, 20B+, and 70B+. 

On the other hand, we also evaluate some prominent close-sourced LLMs including GPT-4o, OpenAI-o1, Claude, GLM4, DeepSeek-v2.5, Qwen-Max,  and the upcoming Doubao-Coder-Preview. The access links of the open-sourced and close-sourced models are listed in Table~\ref{tab:open_source_model} and Table~\ref{tab:api_model}, respectively.


\paragraph{Implementation Details.}
For open-sourced coders, we pull model checkpoints from Hugging Face\footnote{\url{https://huggingface.co/}} and load the model with \texttt{vLLM}~\citep{vllm} to accelerate the evaluation. We set \textit{temperature} to 0, \textit{max\_completion\_tokens} to 2048, and keep all other settings as default. For the input prompt, we set each problem text in \benchmark{} as the user prompt, while leaving the system prompt as default for each model.

We chat with the close-sourced models via API calls similarly, where \textit{temperature} is set to 0 and \textit{max\_completion\_tokens} is set to 2048. The prompt template is kept the same.
After model inference, the first code block with the corresponding programming language formatted in Markdown is extracted from the generated output. If no Markdown code block is detected, a heuristic approach is employed to identify and extract incomplete code snippets. The extracted code, combined with predefined test cases, is then used to synthesize the complete code, which is evaluated for correctness.

\begin{figure}[th]
	\centering
    \begin{minipage}[c]{0.4\textwidth}
        \centering
        \includegraphics[width=\textwidth]{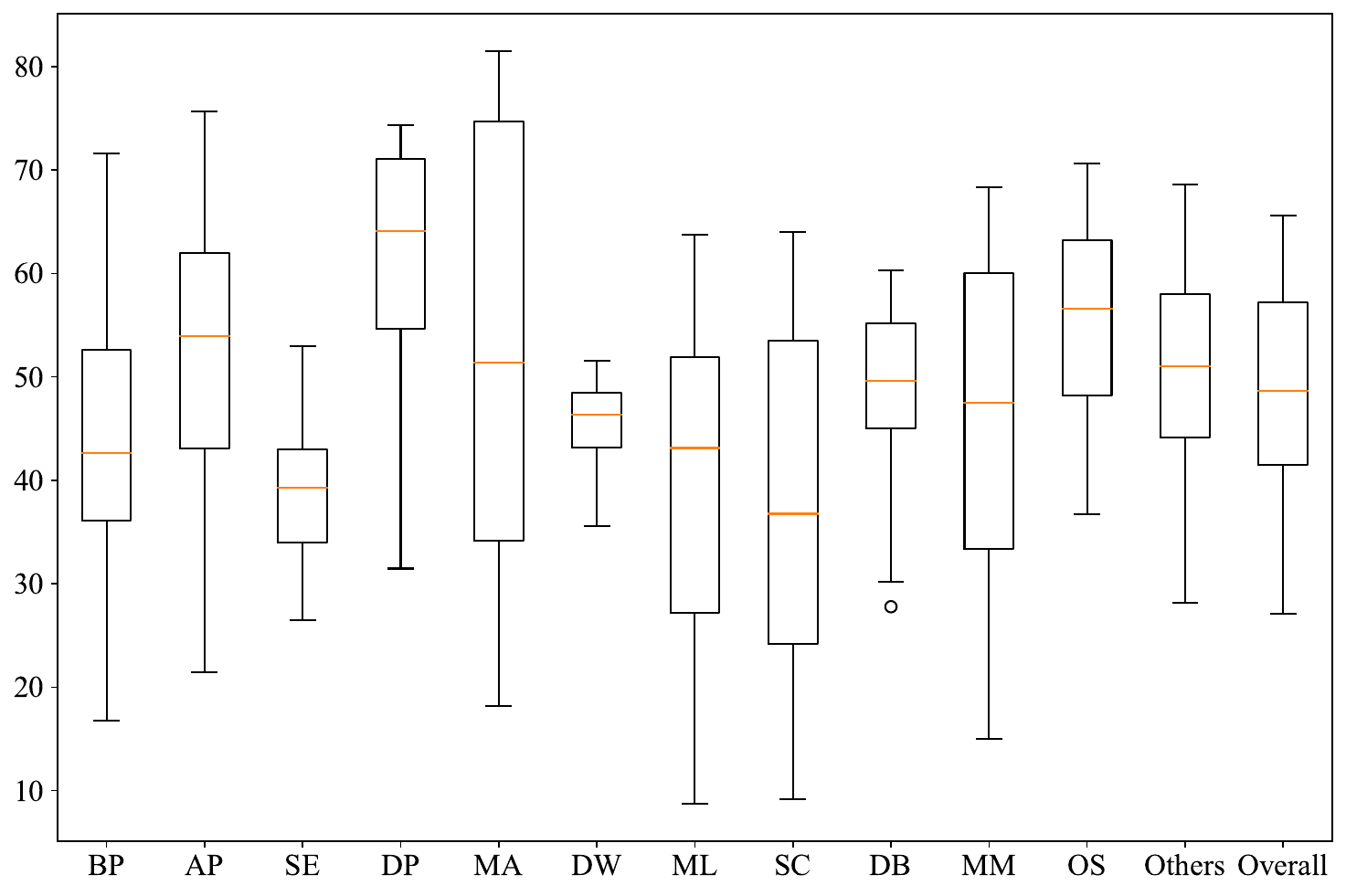}
        \subcaption{General performance}
        \label{fig:category_box}
    \end{minipage}
    \begin{minipage}[c]{0.4\textwidth}
        \centering     
        \includegraphics[width=\textwidth]{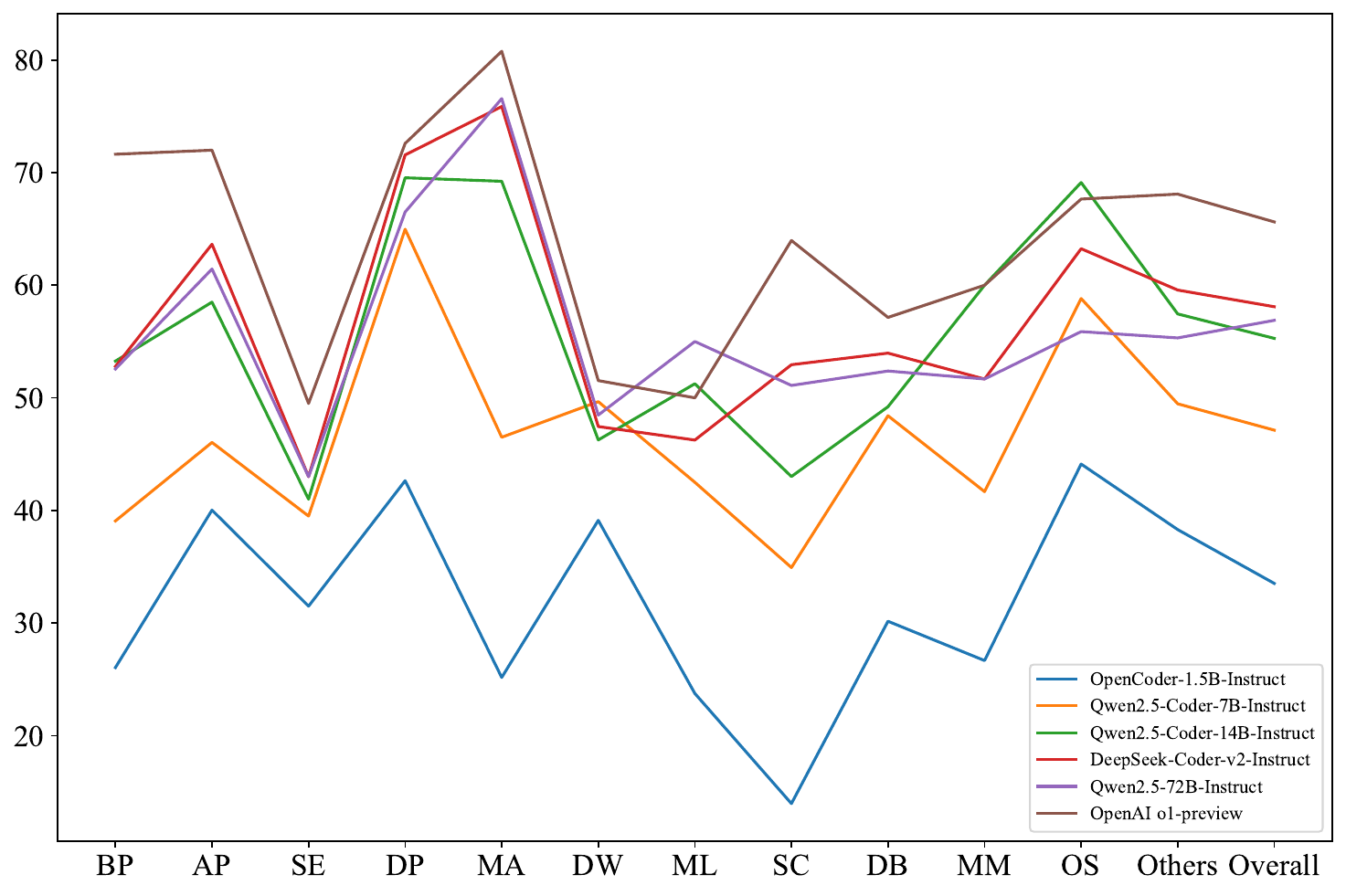}
        \subcaption{Performance of selected models}
        \label{fig:category_line}
    \end{minipage}
	\caption{\label{fig:category} Visualization on domain performance.}
\end{figure}
\subsection{Results and Analysis}
We conduct a systematic evaluation of those AI coders with \benchmark. Results across the 11+ real-world domains are presented in Table~\ref{tab:category}. Owing to the powerful reasoning capability, OpenAI o1-preview unsurprisingly leads the board. However, the dominant position of closed-sourced models has been challenged, with some closed-sourced models being matched or even surpassed by pioneers in open-sourced ones. DeepSeekCoder-v2-Instruct, a 236B-MoE model, is the best behavior of open-sourced models, which pulls away the runner-ups in AP, OS, and Others. OpenCoder-1.5B-Instruct, Qwen2.5-Coder-7B-Instruct, and Qwen2.5-Coder-14B-Instruct achieve the top spot in their groups and outperform some models in the closest higher level.

As illustrated in Figure~\ref{fig:category}, we visualize the model performance on different domains in \benchmark. From Figure~\ref{fig:category_box}, we could find that the performance of AI coders varies significantly in BP, AP, MA, ML, SC, and MM. The largest range occurs in MA, with the best mathematician being OpenAI o1-mini (81.47) while the worst is CodeLlama-34B-Instruct (18.18). Mathematical programming requires models to be proficient in both math and code, and those trained on a code-highly-concentrated corpus would struggle to achieve high scores in MA. Similarly, the variances of ML, SC, and MM, are also remarkable, as each of these problems requires domain knowledge beyond coding.
Moreover,
in Figure~\ref{fig:category_line}, we visualize the first place in each model division. The performance trends of OpenAI-o1-preview, Qwen2.5-72B-Instruct, DeepSeek-Coder-v2-Instruct, and Qwen2.5-Coder-14B-Instruct are generally consistent. OpenAI-o1-Preview has a clear advantage on BP, AP, and SC. On ML and OS, OpenAI-o1-preview only achieves the second place. On the remaining domains, there are slight performance gaps between OpenAI-o1-preview and the runner-ups. Independent of those larger models, the performance trends of OpenCoder-1.5B-Instruct and Qwen2.5-Coder-7B-Instruct are close to each other. These two small models fall behind on BP, AP, MA, and SC, suggesting that reasoning power in these domains may only emerge in the larger models.

\begin{table}[t]
\resizebox{0.98\textwidth}{!}{
\begin{tabular}{l|ccccccccccccccccc}
\toprule[1.5pt]
\multicolumn{1}{c}{Model}       & Bash  & C++   & C\#   & D     & Go    & HTML  & Java  & JS    & PHP   & Python & R     & Ruby  & Rust           & Scala & SQL   & TS    & Overall        \\
\midrule
\multicolumn{18}{c}{1B+ Instruction Tuned Coder} \\
\midrule
OpenCoder-1.5B-Instruct &
73.33 & 20.09 & 50.00 & 10.87 & 39.71 & 51.88 & 38.65 & 51.28 & 58.33 & 29.27 & 20.00 & 42.11 & 33.33 & 28.57 & 35.00 & 50.00 & \textbf{33.52} \\
Qwen2.5-Coder-1.5B-Instruct &
86.67 & 12.62 & 22.22 & 6.52 & 42.65 & 58.75 & 37.99 & 51.28 & 47.22 & 29.09 & 12.50 & 21.05 & 16.67 & 7.14 & 43.75 & 57.89 & 30.74 \\
DeepSeek-Coder-1.3B-Instruct    & 60.00 & 10.28 & 43.94 & 4.35 & 30.88 & 46.88 & 35.15 & 52.56 & 63.89 & 24.35 & 5.00 & 34.21 & 15.00 & 7.14 & 28.75 & 44.74 & 27.65 \\
\midrule
\multicolumn{18}{c}{6B+ Instruction Tuned Coder}  \\
\midrule
Qwen2.5-Coder-7B-Instruct       & 
93.33 & 34.58 & 58.08 & 23.91 & 54.41 & 65.63 & 43.01 & 73.08 & 63.89 & 47.22 & 18.75 & 52.63 & 56.67 & 35.71 & 60.00 & 68.42 & \textbf{47.95}  \\
Yi-Coder-9B-Chat                
& 76.67 & 25.23 & 75.25 & 13.04 & 41.18 & 67.50 & 42.58 & 74.36 & 55.56 & 45.62 & 27.50 & 63.16 & 51.67 & 46.43 & 56.25 & 65.79 & 47.13 \\
OpenCoder-8B-Instruct           
& 66.67 & 29.91 & 61.62 & 30.43 & 48.53 & 61.25 & 41.27 & 70.51 & 58.33 & 40.05 & 25.00 & 39.47 & 50.00 & 51.79 & 61.25 & 60.53 & 43.63 \\
DeepSeek-Coder-7B-Instruct-v1.5 
& 50.00 & 24.77 & 68.18 & 28.26 & 54.41 & 63.13 & 40.83 & 66.67 & 44.44 & 41.29 & 21.25 & 52.63 & 43.33 & 39.29 & 55.00 & 50.00 & 43.48          \\
DeepSeek-Coder-6.7B-Instruct    
& 76.67 & 23.36 & 56.06 & 13.04 & 52.94 & 65.00 & 41.48 & 65.38 & 50.00 & 39.34 & 31.25 & 47.37 & 43.33 & 28.57 & 57.50 & 60.53 & 41.88          \\
CodeQwen1.5-7B-Chat             
& 73.33 & 30.37 & 41.92 & 15.22 & 48.53 & 57.50 & 44.76 & 60.26 & 44.44 & 37.44 & 15.00 & 52.63 & 58.33 & 46.43 & 52.50 & 60.53 & 40.52          \\
CodeLlama-7B-Instruct           
& 76.67 & 13.08 & 43.94 & 11.96 & 23.53 & 54.38 & 32.97 & 37.18 & 27.78 & 21.98 & 17.50 & 21.05 & 20.00 & 25.00 & 41.25 & 50.00 & 27.06          \\
\midrule
\multicolumn{18}{c}{13B+ Instruction Tuned Coder} \\
\midrule
Qwen2.5-Coder-14B-Instruct &
93.33 & 39.25 & 65.66 & 41.30 & 63.24 & 66.25 & 43.45 & 80.77 & 77.78 & 56.22 & 32.50 & 71.05 & 73.33 & 42.86 & 62.50 & 68.42 & \textbf{55.28} \\
DeepSeekCoder-v2-Lite-Instruct  & 50.00 & 36.45 & 52.53 & 27.17 & 61.76 & 64.38 & 41.27 & 73.08 & 58.33 & 49.35 & 38.75 & 55.26 & 53.33 & 42.86 & 57.50 & 60.53 & 48.73          \\
StarCoder2-15B-Instruct-v0.1    
& 56.67 & 21.03 & 60.61 & 29.35 & 47.06 & 49.38 & 31.44 & 70.51 & 44.44 & 42.36 & 28.75 & 71.05 & 35.00 & 32.14 & 63.75 & 52.63 & 41.79 \\
CodeLlama-13B-Instruct          
& 63.33 & 11.68 & 50.00 & 15.22 & 33.82 & 55.63 & 33.84 & 35.90 & 36.11 & 21.33 & 17.50 & 42.11 & 25.00 & 21.43 & 38.75 & 47.37 & 27.59 \\
\midrule
\multicolumn{18}{c}{20B+ Instruction Tuned Coder} \\
\midrule
DeepSeekCoder-v2-Instruct       & 83.33 & 43.46 & 72.73 & 28.26 & 66.18 & 69.38 & 42.36 & 80.77 & 61.11 & 60.43 & 41.25 & 65.79 & 68.33 & 69.64 & 66.25 & 68.42 & \textbf{58.09} \\
Qwen2.5-Coder-32B-Instruct &
83.33 & 36.92 & 76.77 & 46.74 & 54.41 & 71.25 & 40.39 & 79.49 & 58.33 & 59.06 & 35.00 & 63.16 & 76.67 & 50.00 & 70.00 & 57.89 & 56.88 \\
DeepSeekCoder-33B-Instruct      & 60.00 & 26.64 & 68.18 & 19.57 & 57.35 & 71.25 & 44.10 & 66.67 & 50.00 & 48.10 & 32.50 & 60.53 & 50.00 & 46.43 & 60.00 & 57.89 & 48.61          \\
CodeLlama-34B-Instruct          & 76.67 & 12.62 & 47.98 & 11.96 & 35.29 & 63.75 & 33.62 & 37.18 & 44.44 & 23.34 & 18.75 & 36.84 & 21.67 & 21.43 & 48.75 & 47.37 & 29.22          \\
\midrule
\multicolumn{18}{c}{70B+ Instruction Tuned General Language Model} \\
\midrule
Qwen2.5-72B-Instruct            
& 80.00 & 32.24 & 57.07 & 33.70 & 63.24 & 73.75 & 41.92 & 79.49 & 77.78 & 61.02 & 37.50 & 63.16 & 61.67 & 66.07 & 68.75 & 68.42 & \textbf{56.88} \\
Llama3.1-70B-Instruct           
& 76.67 & 28.50 & 72.73 & 40.22 & 50.00 & 77.50 & 28.38 & 71.79 & 41.67 & 54.50 & 36.25 & 65.79 & 58.33 & 46.43 & 68.75 & 57.89 & 51.45 \\
\midrule
\multicolumn{18}{c}{Close-Sourced API Model} \\
\midrule
OpenAI o1-preview &              
90.00 & 51.40 & 80.81 & 55.43 & 64.71 & 75.63 & 42.36 & 78.21 & 75.00 & 68.19 & 62.50 & 84.21 & 86.67 & 83.93 & 71.25 & 78.95 & \textbf{65.62} \\
OpenAI o1-mini                  
& 90.00 & 66.36 & 76.77 & 53.26 & 67.65 & 75.00 & 34.06 & 80.77 & 88.89 & 67.71 & 58.75 & 78.95 & 83.33 & 82.14 & 68.75 & 68.42 & 64.73          \\
Claude-35-Sonnet                
& 86.67 & 41.12 & 75.76 & 64.13 & 50.00 & 81.88 & 40.39 & 70.51 & 47.22 & 66.59 & 46.25 & 84.21 & 80.00 & 83.93 & 71.25 & 55.26 & 62.57          \\
GPT 4o-0806                     
& 93.33 & 50.47 & 72.73 & 28.26 & 67.65 & 71.25 & 43.23 & 83.33 & 44.44 & 64.81 & 52.50 & 73.68 & 85.00 & 73.21 & 76.25 & 57.89 & 61.77          \\
Doubao-Coder-Preview
& 76.67 & 40.65 & 74.24 & 43.48 & 52.94 & 66.25 & 37.77 & 84.62 & 77.78 & 62.74 & 45.00 & 71.05 & 66.67 & 75.00 & 66.25 & 65.79 & 58.92              \\
DeepSeek-v2.5                   
& 86.67 & 45.33 & 74.75 & 30.43 & 61.76 & 76.25 & 40.83 & 75.64 & 91.67 & 60.90 & 35.00 & 68.42 & 65.00 & 64.29 & 66.25 & 71.05 & 58.65          \\
GLM-4-Plus                       
& 83.33 & 38.79 & 74.24 & 32.61 & 54.41 & 75.63 & 41.70 & 80.77 & 80.56 & 58.29 & 31.25 & 52.63 & 60.00 & 58.93 & 72.50 & 55.26 & 56.40          \\
Qwen-Max                        
& 76.67 & 32.71 & 40.40 & 28.26 & 61.76 & 73.75 & 42.79 & 85.90 & 52.78 & 60.37 & 35.00 & 60.53 & 76.67 & 46.43 & 72.50 & 52.63 & 55.16   \\      
\bottomrule
\end{tabular}
}
\caption{\label{tab:program_lang} Model performance across programming languages.}
\end{table}

\begin{figure}[th]
	\centering
	\includegraphics[width=0.6\textwidth]{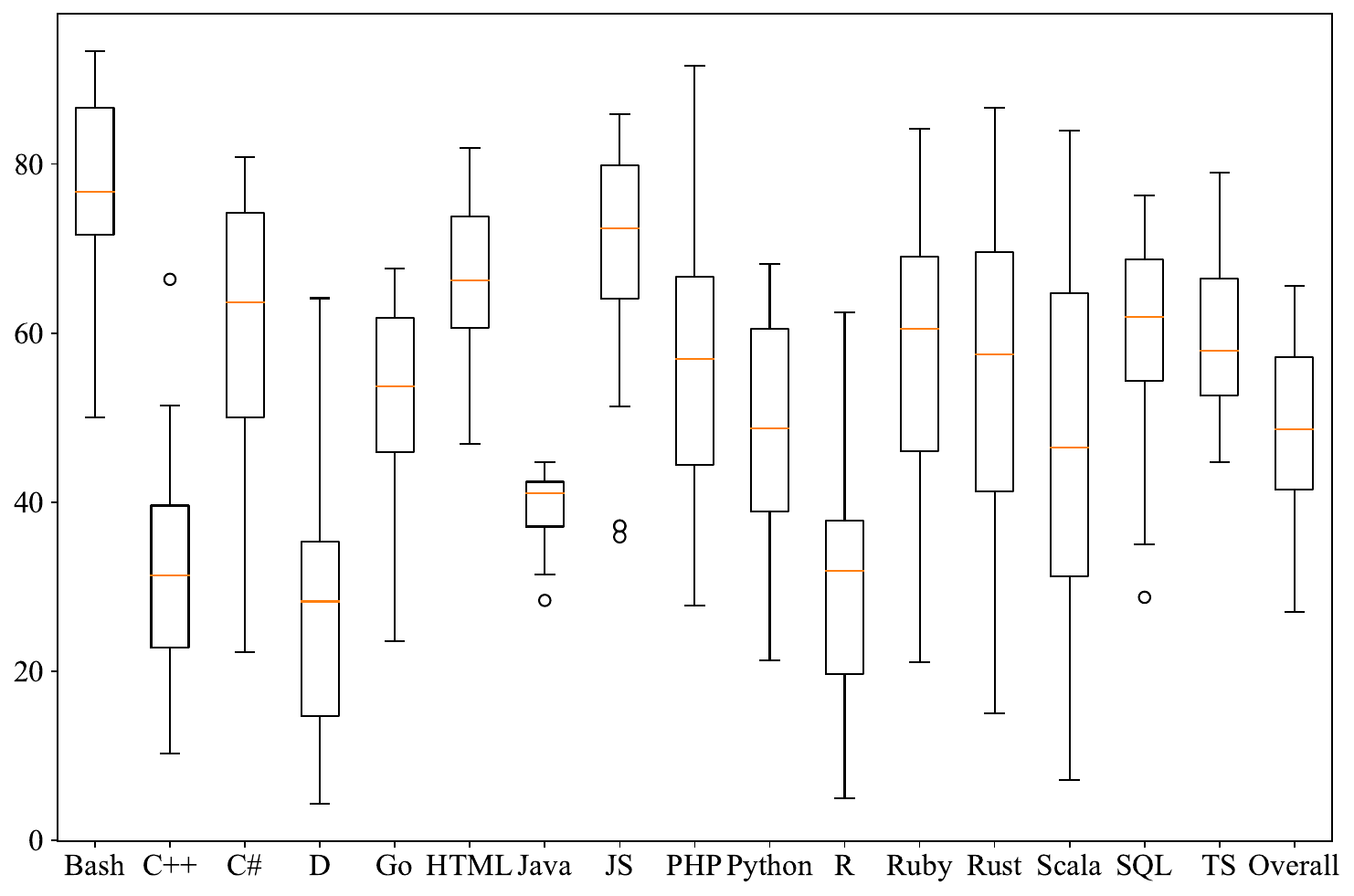}
	\caption{\label{fig:program} General performance on different programming languages.}
\end{figure}

\begin{figure}[th]
	\centering
    \begin{minipage}[c]{0.45\textwidth}
        \centering
        \includegraphics[width=\textwidth]{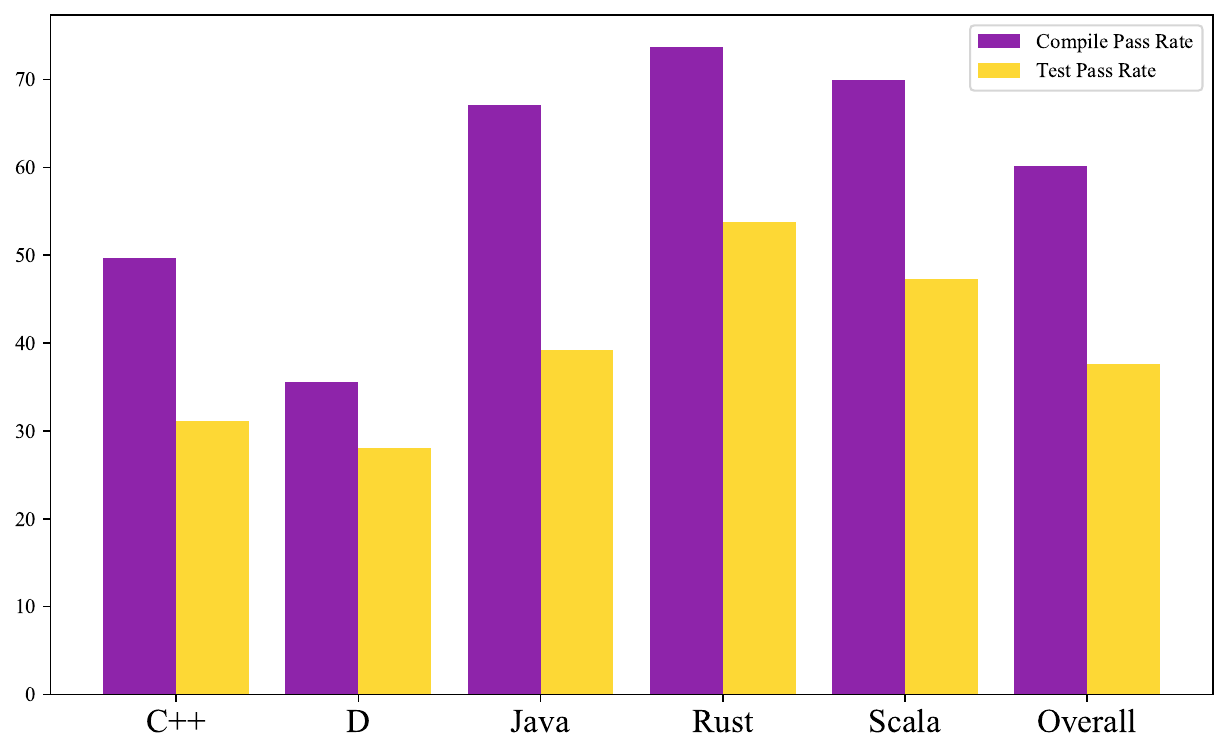}
        \subcaption{Compile Pass@1 and Test Pass@1}
        \label{fig:compile_rate}
    \end{minipage}
    \begin{minipage}[c]{0.45\textwidth}
        \centering     
        \includegraphics[width=\textwidth]{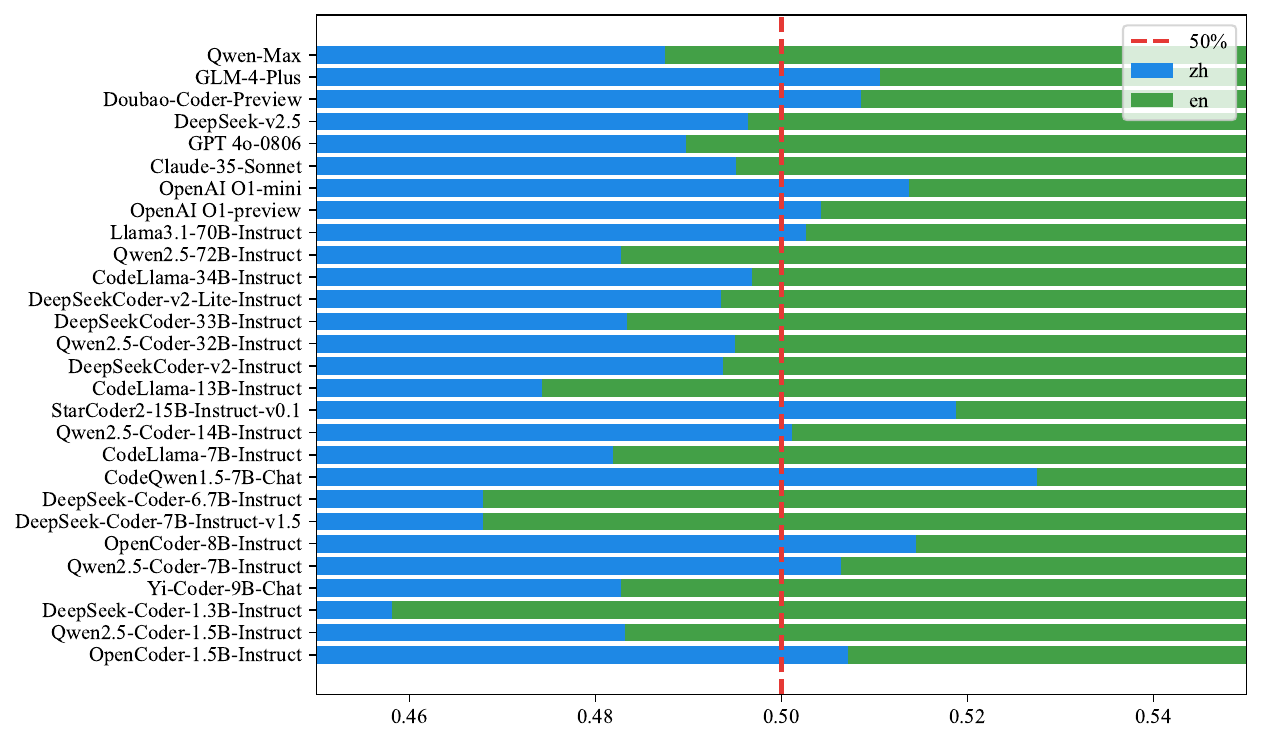}
        \subcaption{Pass@1 in Chinese/English.}
        \label{fig:enzh}
    \end{minipage}
	\caption{\label{fig:compile_and_enzh} Visualization of performance on different languages.}
    \vspace{-3mm}
\end{figure}

\subsection{Analysis on the performance of different programming languages}
We present results of different programming languages in Table~\ref{tab:program_lang} and Figure~\ref{fig:program}. Except for Java, the performance of AI coders on different languages has significant variance. The performance gaps are relatively larger in C\#, D, PHP, Ruby, Rust, and Scala. Besides, as our SandboxFusion provides feedback from the compilers, we evaluate the compilation pass rates of model responses in compiled languages including C++, D, Java, Rust, and Scala. In Figure~\ref{fig:compile_rate}, there is a positive correlation between the compile pass rate and the test pass rate, but a compile pass does not necessarily mean a test pass.
Moreover,
from Figure~\ref{fig:enzh}, we found that the written language of the prompts would affect the model performance. We are surprised that DeepSeek-Coder-1.5B-Instruct, proposed by a Chinese institution, is the most favorable model for English questions. On the other hand, some native English speakers, such as StarCoder2-15B-Instruct-v0.1, Open AI-o1-preview, and Open AI-o1-mini, perform better on Chinese questions.

\subsection{Scaling Laws on \benchmark{}}
\label{sec:scaling}
\begin{wrapfigure}{r}{0.45\textwidth}
  \centering
  \includegraphics[width=0.45\textwidth]{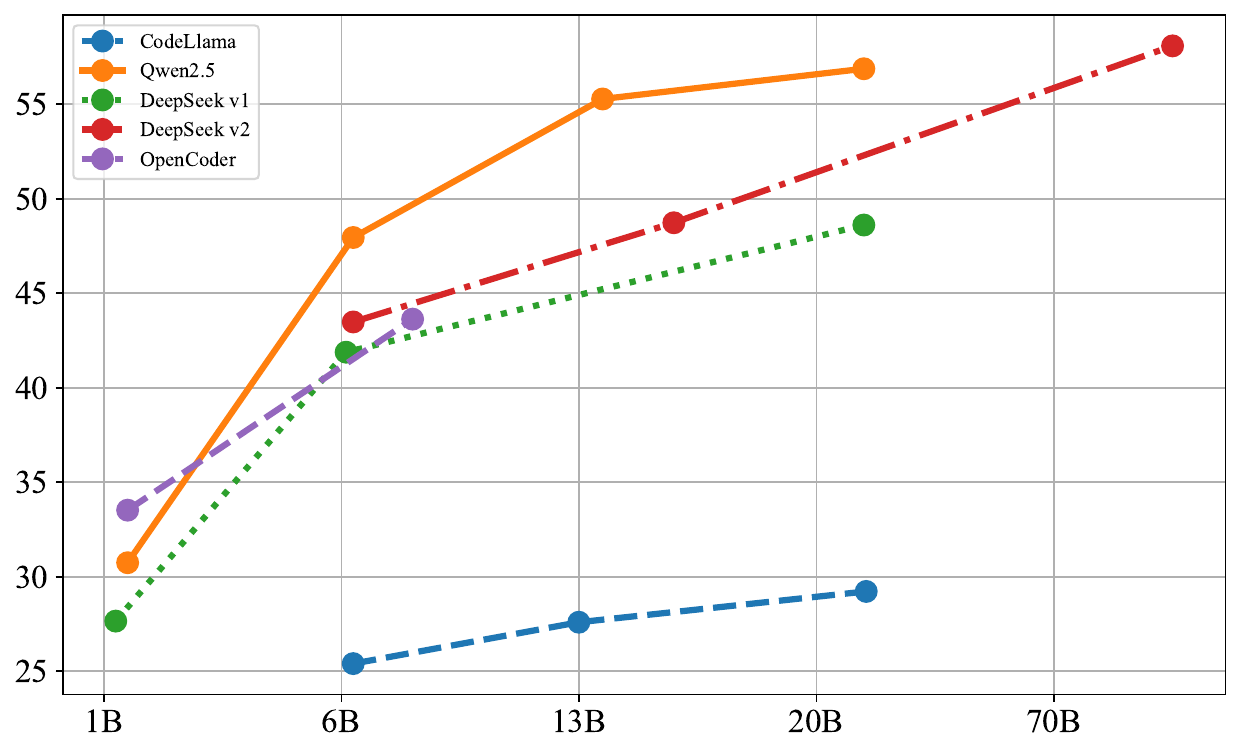}
  \vspace{-3mm}
  \caption{Performance of different sizes.}
  \label{fig:scaling}
  \vspace{-8mm}
\end{wrapfigure}
We categorized the model into 5 series based on the criteria in Table~\ref{tab:series} of Appendix and visualize the performance of different model series in Figure~\ref{fig:scaling}. As the parameter increases, performance gains are achieved for all the model families. We can say that the scaling law still holds, but the improvement in model performance is diminishing as the model size increases.

\subsection{Analysis on the performance of different difficulties}
Figure~\ref{fig:difficulty} presents model performance on different difficulties. As stated in Section~\ref{sec:data_quality}, we implement a voting method with the involvement of six AI coders to determine the difficulty of each problem. Overall, the 1B+ models and CodeLlama series are less effective on all difficulty levels. While the rest of the models could solve simple problems equally well, and gaps appear in the medium questions. As for hard questions, the closed-source models generally outperform the open-source coders.

\subsection{Analysis on the effect of feedback from SandboxFusion}
As shown in Figure~\ref{fig:reflection}, to demonstrate the effectiveness of the feedback using SandboxFusion,
we compare the ``Reflection'' and ``BoN'' strategies.
For ``Reflection'', we reproduce the self-refine strategy~\citep{madaan2024self} by refining the answers of $N$ times using the feedback context of SandboxFusion. 
For ``BoN'', we just infer $N$ times to obtain the results.
In Figure~\ref{fig:reflection},
we observe that the ``Reflection'' is better than ``BoN'' a lot,
which demonstrates the effectiveness of the feedback context provided by the SandboxFusion.
\begin{figure}[t]
	\centering
    \includegraphics[width=0.85\textwidth]{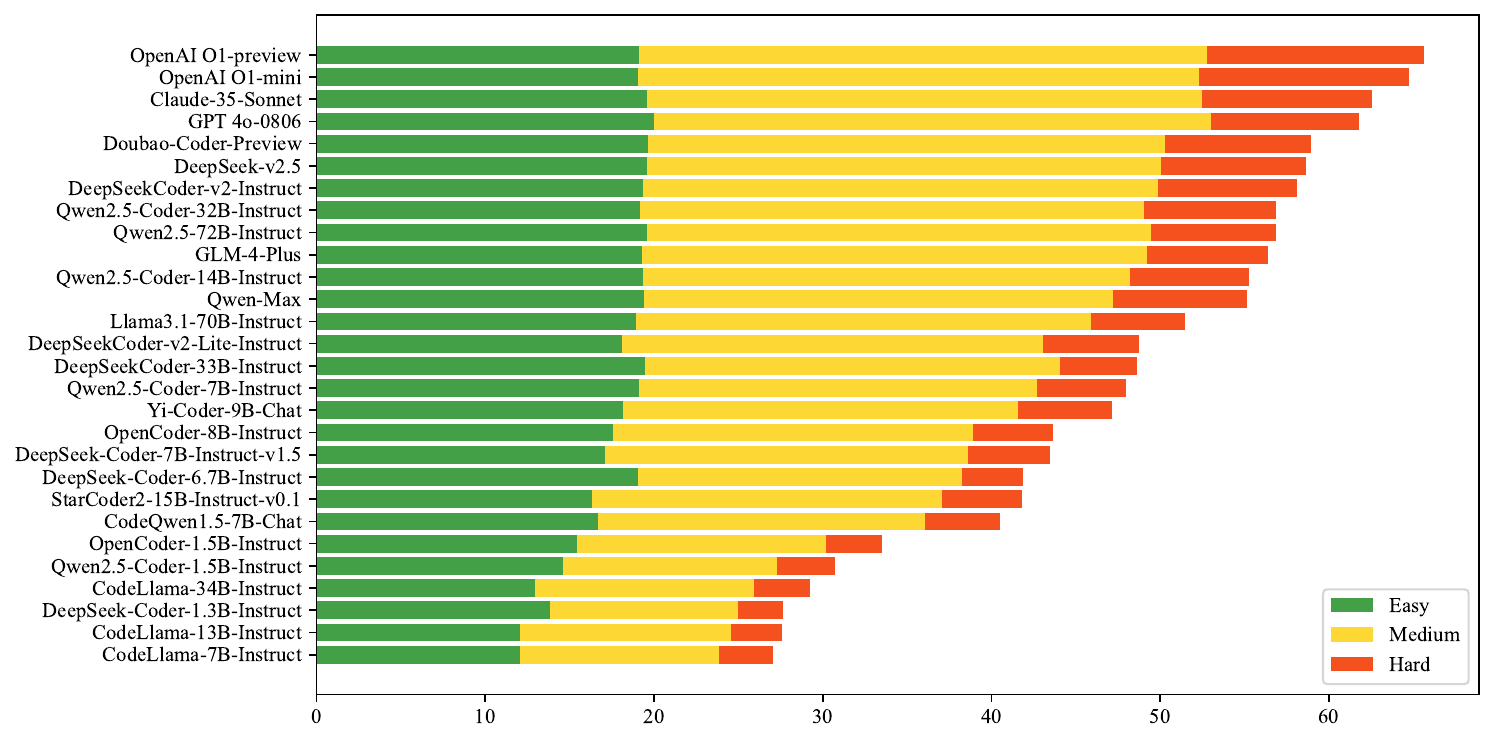}
	\caption{\label{fig:difficulty} Visualization of performance on different difficulties.}
    \vspace{-3mm}
\end{figure}
\begin{figure}[t]
    \centering
    \includegraphics[width=0.5\linewidth]{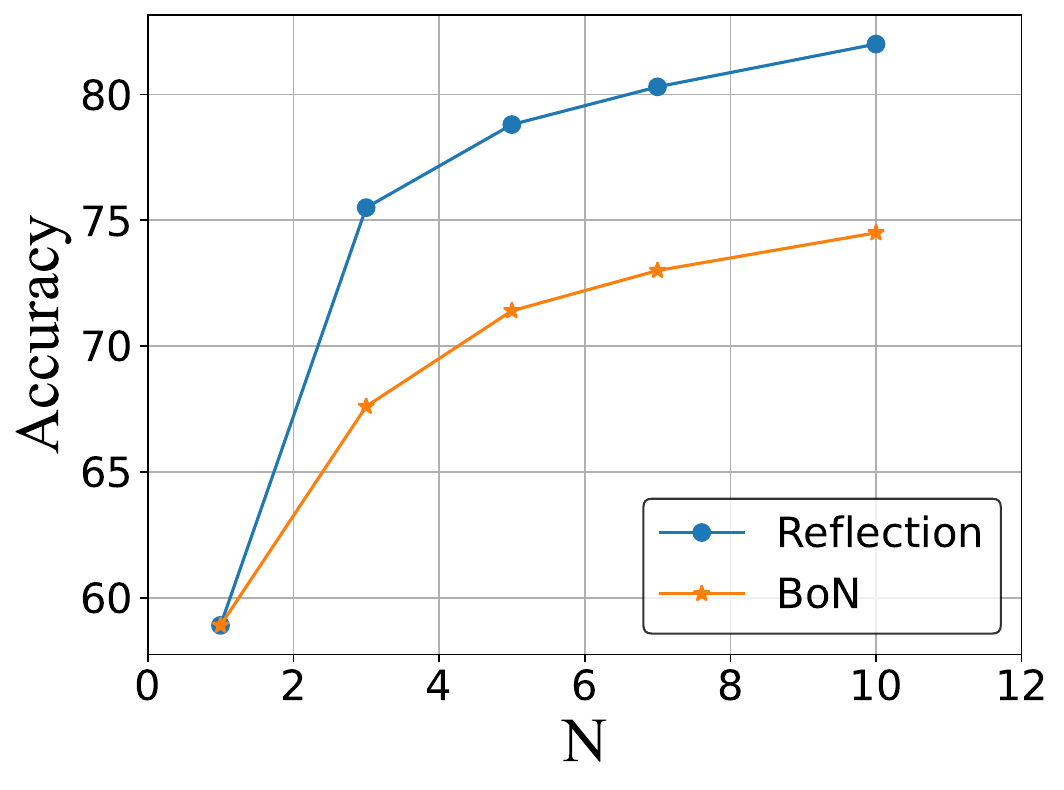}
    \caption{Comparison between BoN and Reflection.}
    \label{fig:reflection}
\end{figure} 

\section{Related Works}

\paragraph{Code Large Language Models.}
Code large language models (LLMs)~\citep{Chen2021Evaluating, Zhao2024CodeGemmaOC,gpt-neo, black2022gptneox, Le2022CodeRLMC, chowdhery2023palm, nijkamp2022codegen, fried2022incoder, xu2022systematic,unicoder,qwen25coder} has shown powerful capabilities in code generation~\citep{li2022competition,allal2023santacoder}, code debug, code translation~\citep{codegeex,li2023starcoder}, and other coding tasks, which is essential for modern software engineering, For example, there is a wide variety of in-file benchmarks to evaluate different capabilities of code LLMs ~\citep{codegeex,mbpp,livecodebench}, which focus on a limited range of programming languages (e.g. Python and Java). 
Further, recent code LLMs such as Code Llama~\citep{codellama}, DeepSeek-Coder~\citep{deepseek_coder}, and Qwen2.5-Coder~\citep{qwen25coder} gains remarkable progress in multilingual programming code generation and debugging tasks, such as MultiPL-E \citep{multiple}, McEval~\citep{mceval}, and MdEval~\citep{mdevl}.

\noindent\textbf{Code Benchmark.}
Program synthesis is an important task for code LLM, which forces the LLM to read the natural language description and then generates the corresponding code snippet meeting the user requirements~\citep{athiwaratkun2022multi, austin2021program,gu2024cruxeval,lai2023ds1000,liu2023evalplus, yu2024codereval,lu-etal-2022-reacc}.
To further comprehensively evaluate the different aspects of capabilities of LLMs, numerous benchmarks are proposed, such as code translation~\citep{jiao2023evaluation, yan2023codetransocean, zhu2022xlcost}, code retrieval~\citep{huang2021cosqa, husain2019codesearchnet, li2024procqa, lu2021codexglue}, and vulnerability repair~\citep{huq2022review4repair, prenner2023runbugrun, richter2022tssb, tian2024debugbench,mdevl}, and structured data understanding~\citep{tablebench,tablegpt2}.
The recent work McEval~\citep{mceval} extends the number of programming languages to 40 for multilingual evaluation scenarios and MdEval~\citep{mdevl} creates a multilingual code debugging benchmark covering nearly 20 programming languages.
Besides, many benchmarks have been proposed on repository-level code tasks~\citep{agrawal2023guiding, allal2023santacoder, bairi2023codeplan, ding2022cocomic, liu2023repobench, pei2023better, shrivastava2022repository, shrivastava2023repofusion, zhang2023repocoder, Liu2024M2rcEvalMM,Deng2024R2C2CoderEA}. However, most previous workers focus on the capabilities of one aspect of the LLMs, neglecting to test the abilities of the LLM as a program developer across various real-world code development scenarios. In this work, we propose the FullStack Bench to assess the capabilities of LLMs across various real-world code development scenarios

\section{Conclusion}
In this paper, we provide a more holistic evaluation framework \benchmark{} with a corresponding effective execution environment SandboxFusion for code intelligence, which aims to evaluate multilingual programming capabilities
in real-world code development scenarios.
Specifically,
first,
our \benchmark{} mainly involves  mainstream application domains (e.g., basic programming, software engineering, and machine learning ) from 3374 problems,
where each problem has corresponding unit test cases.
Second,
our SandboxFusion  has three distinct features (i.e., Supporting various languages, Easy-to-deploy and Unified multi-dataset execution environment),
which can satisfy the requirements of evaluating \benchmark{}.
Finally, we hope that \benchmark{} could guide the researchers to better understand the code intelligence abilities of existing LLMs and accelerate the growth of foundation models.


\section{Acknowledgements}
We extend our heartfelt gratitude to the larger Seed team, whose dedication and expertise were crucial to the success of this project. Special thanks go to our engineering team for their technical prowess; our data teams, whose diligent efforts in data collection, annotation, and processing were indispensable; our evaluation team for their rigorous testing and insightful feedback; and the Seed-Foundation team for their valuable knowledge sharing. Their contributions have been instrumental to \benchmark{}.

\newpage

\section{Contributions}\label{sec:contributions}

{
\setlength{\parskip}{1em}
\setlength{\parindent}{0em}

\textbf{Siyao Liu} implemented the dataset execution environment together with SandboxFusion, and performed quality assurance.

\textbf{He Zhu} validated the experimental results through analysis and made graphical visualizations. He Zhu, Jiaheng Liu, Jack Yang, Jinxiang Xia, Shukai Liu and Z.Y. Peng are affiliated with M-A-P.

\textbf{Shulin Xin} conducted experiments on pass rate and reflection.

\textbf{Aoyan Li and Yifan Sun} proposed and implemented the framework for domain classification.

\textbf{Jiaheng Liu, Siyao Liu and He Zhu} wrote the paper.

\textbf{Li Chen, Aoyan Li, Siyao Liu, Rui Long, Yongsheng Xiao and Shulin Xin} contributed to dataset samples, established and maintained an iterative process for organizing and curating datasets.

\textbf{Shukai Liu, Z.Y. Peng, Jinxiang Xia and Jack Yang} contributed supplementary samples in minority languages and domains, and performed revisions on these samples.

\textbf{Kai Shen, Liang Xiang and Hongxia Yang} provided research guidance and management support.

\textbf{Yao Cheng, Jianfeng Chen, Jie Chen, Liyu Chen, Wentao Chen, Zhengyu Chen, Shijie Geng, Bo Li, Bowen Li, Linyi Li, Boyi Liu, Kaibo Liu, Qi Liu, Tianyi Liu, Tingkai Liu, Yongfei Liu, Guanghan Ning, Jiahao Su, Jing Su, Tao Sun, Yifan Sun, Yunzhe Tao, Guoyin Wang, Siwei Wang, Xuwu Wang, Yite Wang, Xia Xiao, Chenguang Xi, Jingjing Xu, Shikun Xu, Yingxiang Yang, Jianbo Yuan, Jun Zhang, Yufeng Zhang, Yuyu Zhang, Shen Zheng and Ming Zhu} contributed to dataset samples and provided feedback during dataset iterations.

\textbf{Aoyan Li, Qi Liu, Siyao Liu, Jing Mai, Zihan Wang and Shulin Xin} implemented existing open-source datasets in SandboxFusion.

}

\newpage

\bibliography{main.bib}

\newpage

\appendix

\section{Appendix}
\subsection{Visualization on the cases of \benchmark{}}

\begin{figure}[t]
    \centering
    \includegraphics[width=0.4\linewidth]{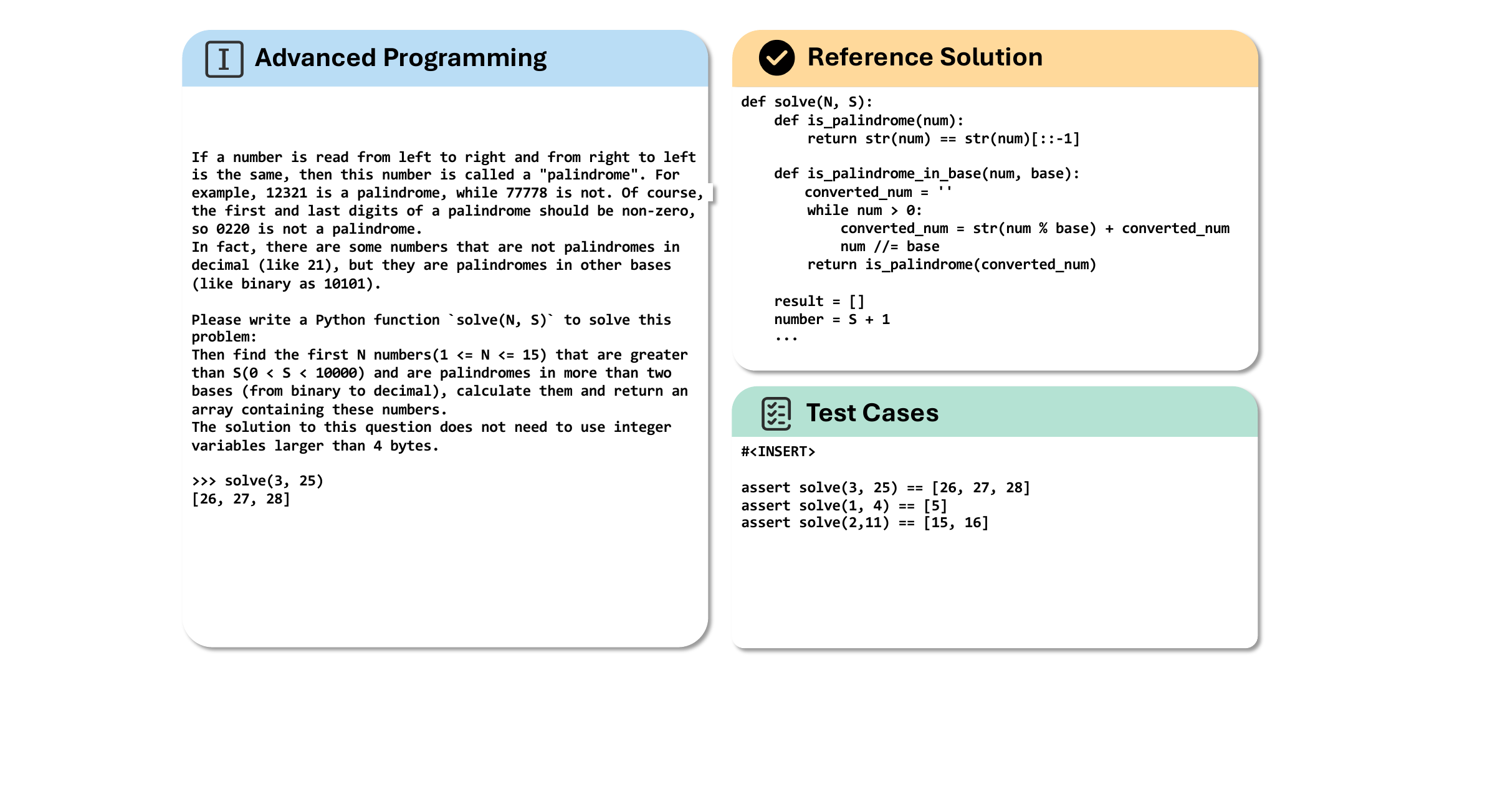} 
    \includegraphics[width=0.40\linewidth]{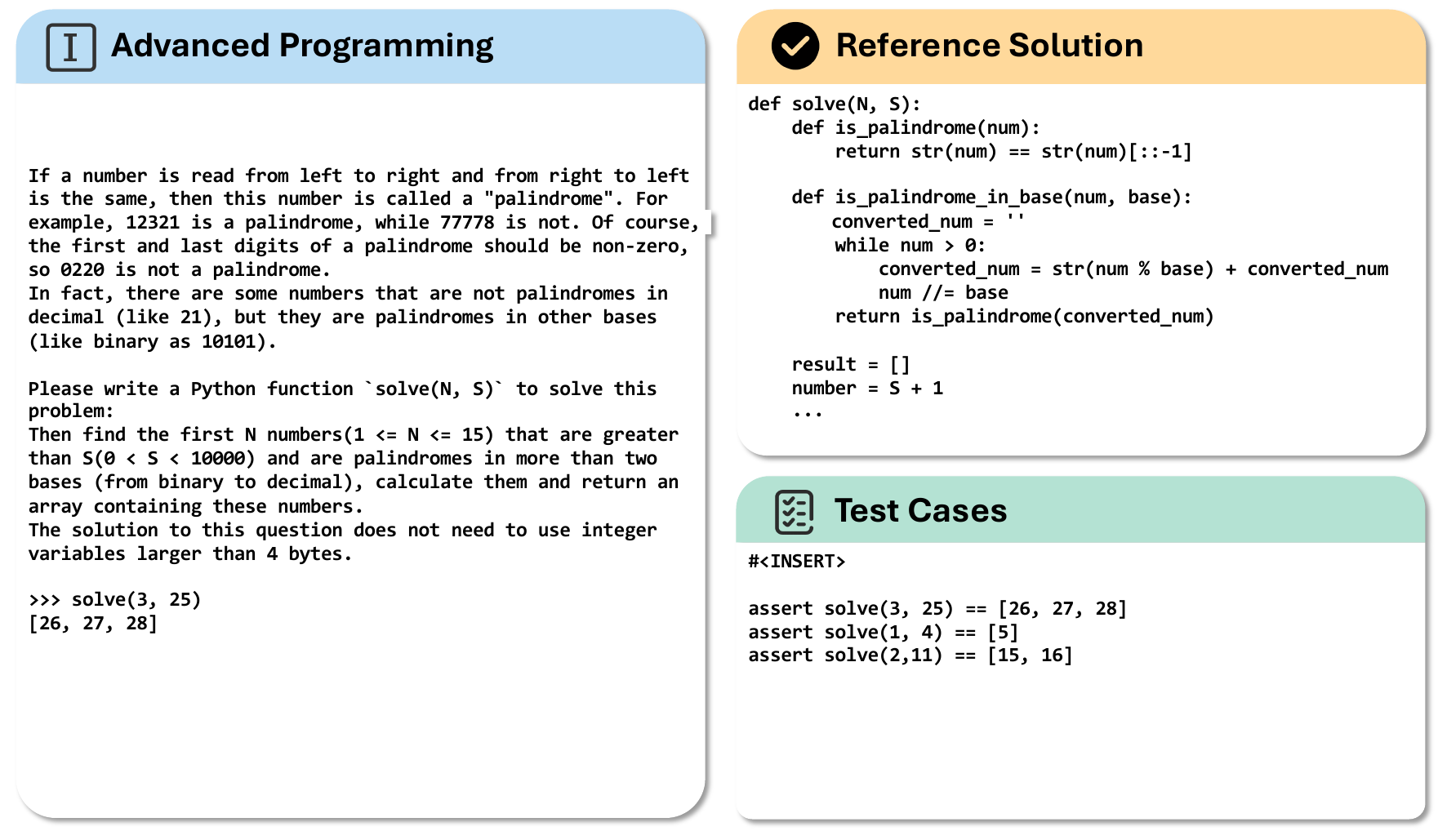}
    \includegraphics[width=0.40\linewidth]{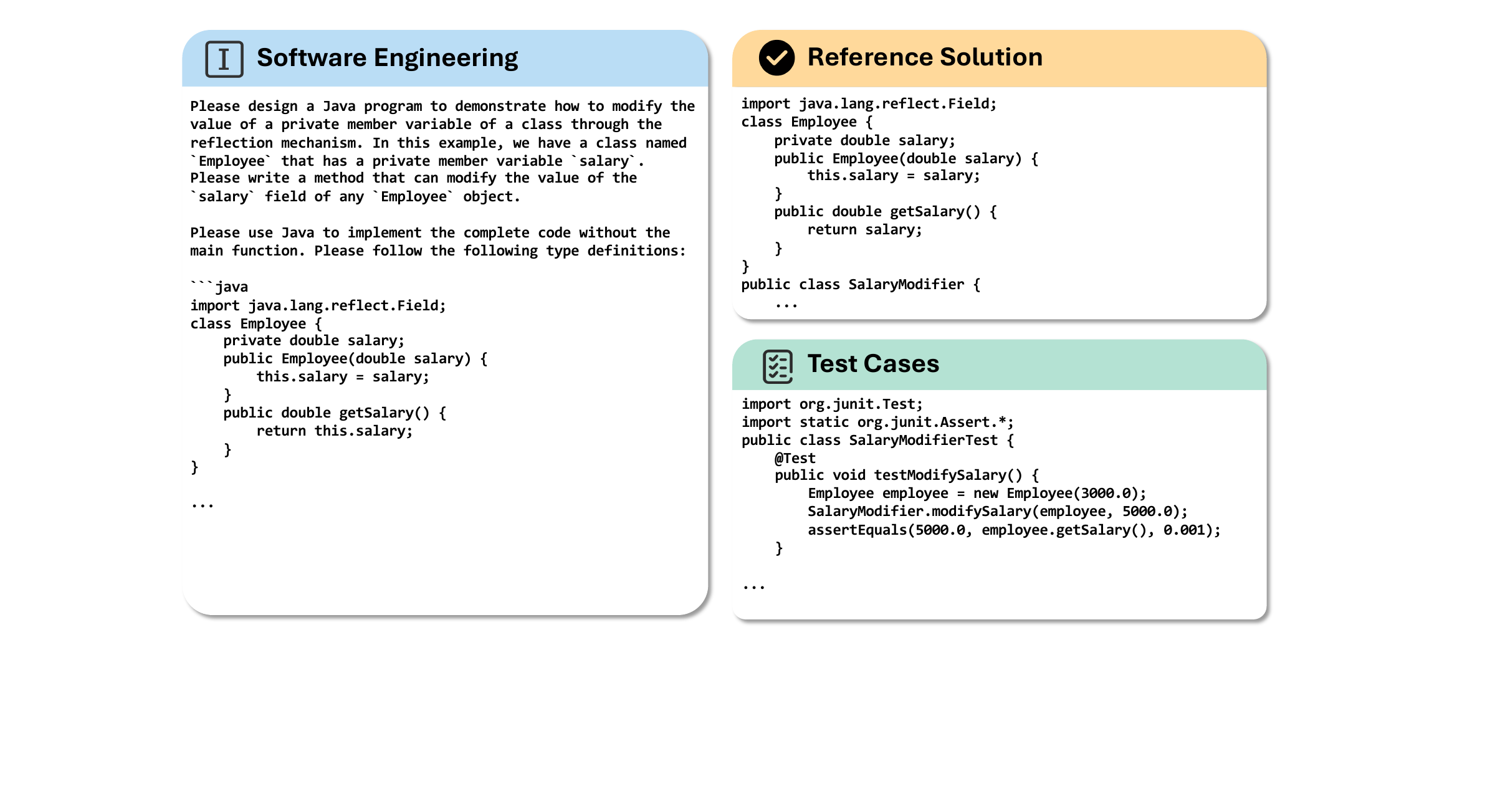}   
            \includegraphics[width=0.40\linewidth]{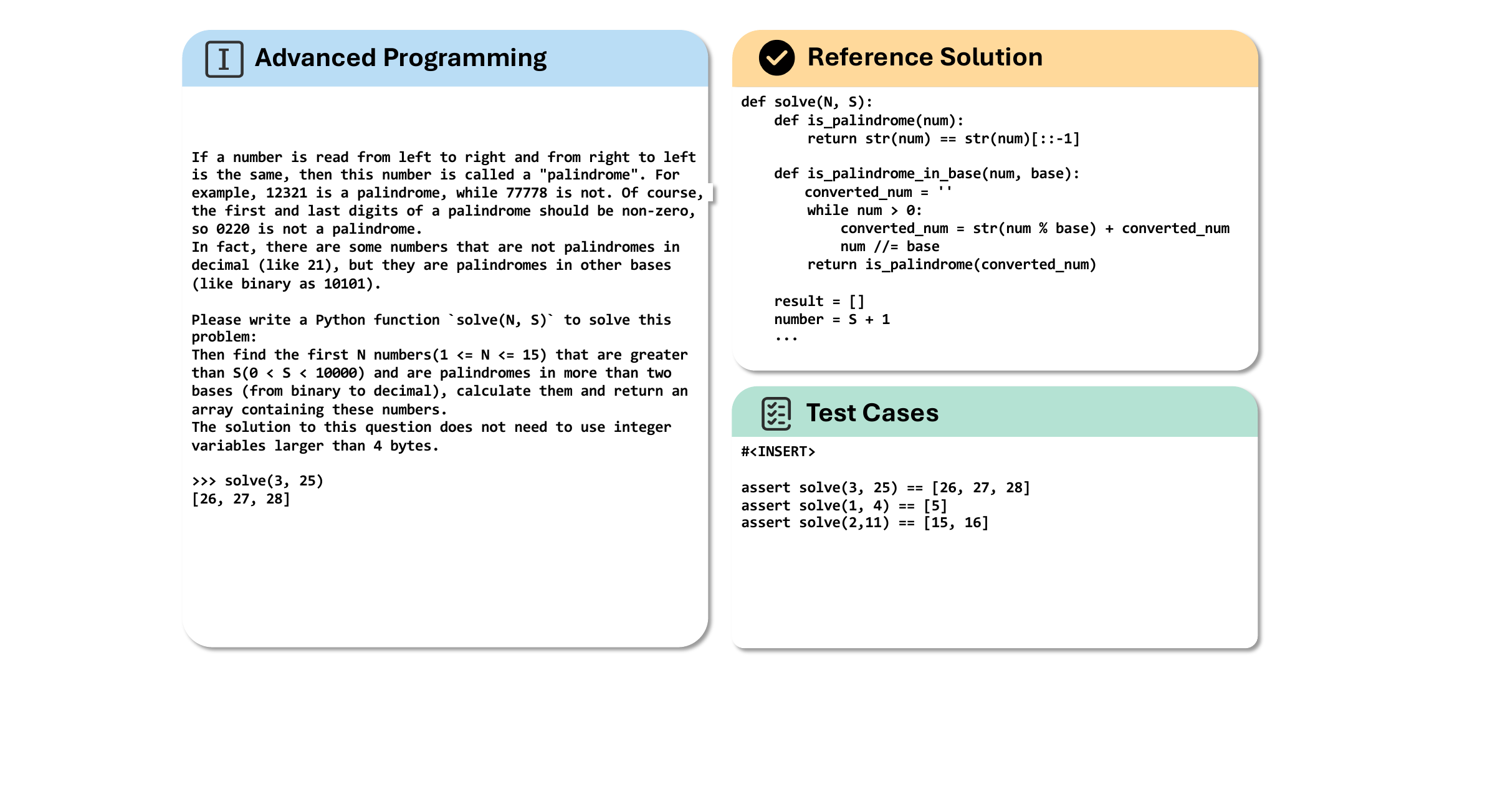}   
              \includegraphics[width=0.40\linewidth]{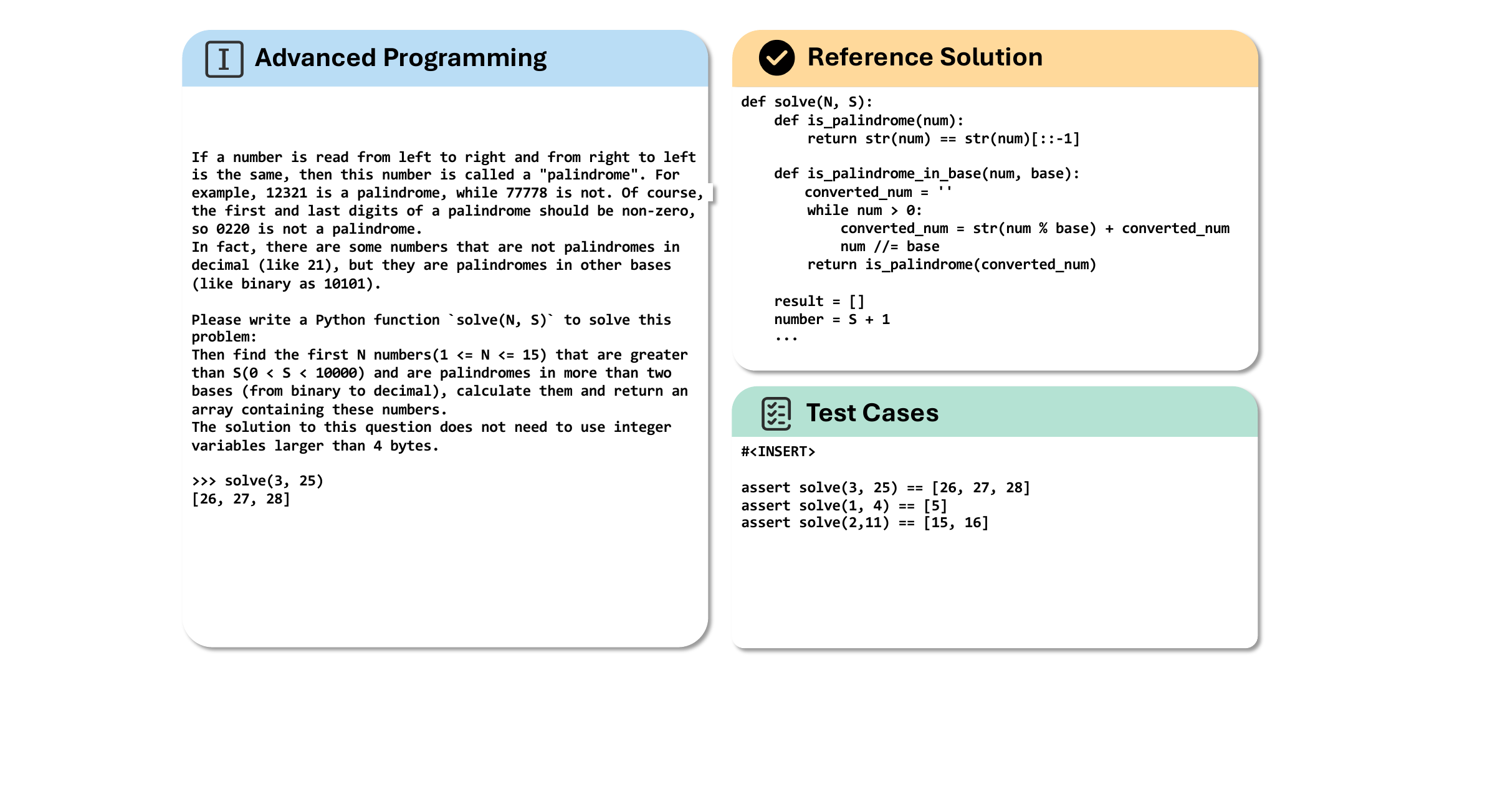} 
            \includegraphics[width=0.40\linewidth]{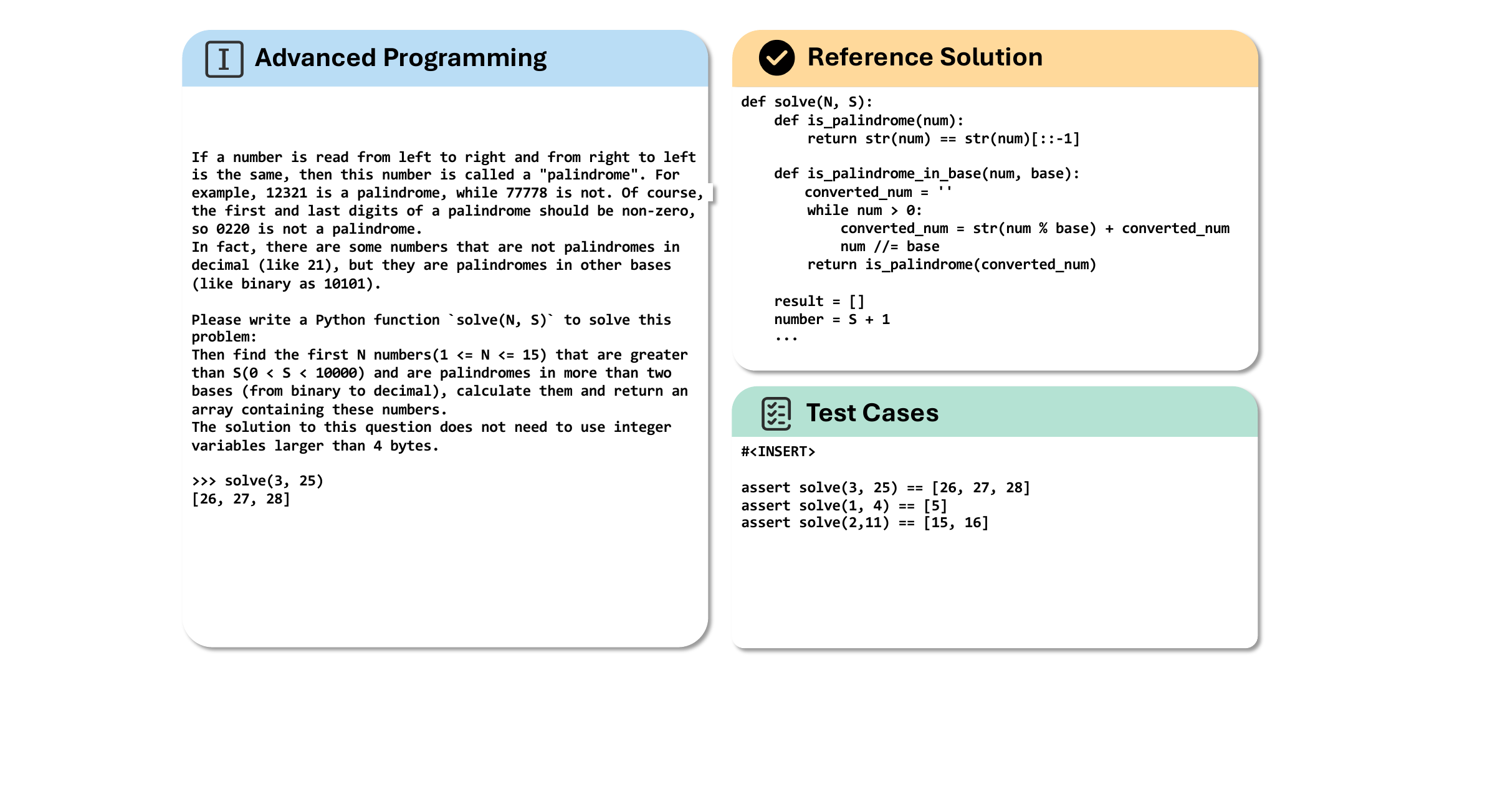}       
            \includegraphics[width=0.40\linewidth]{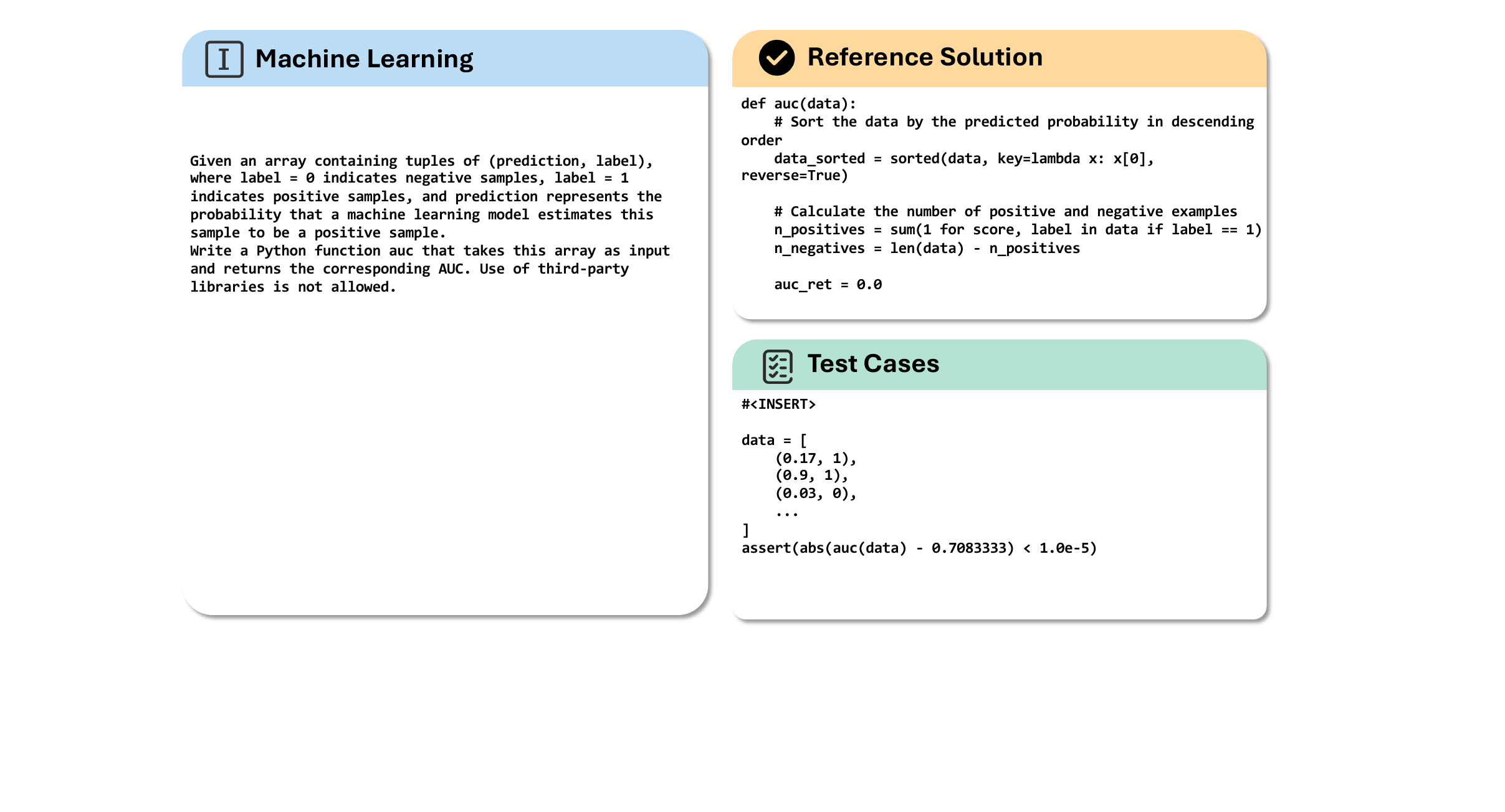} 
                 \includegraphics[width=0.40\linewidth]{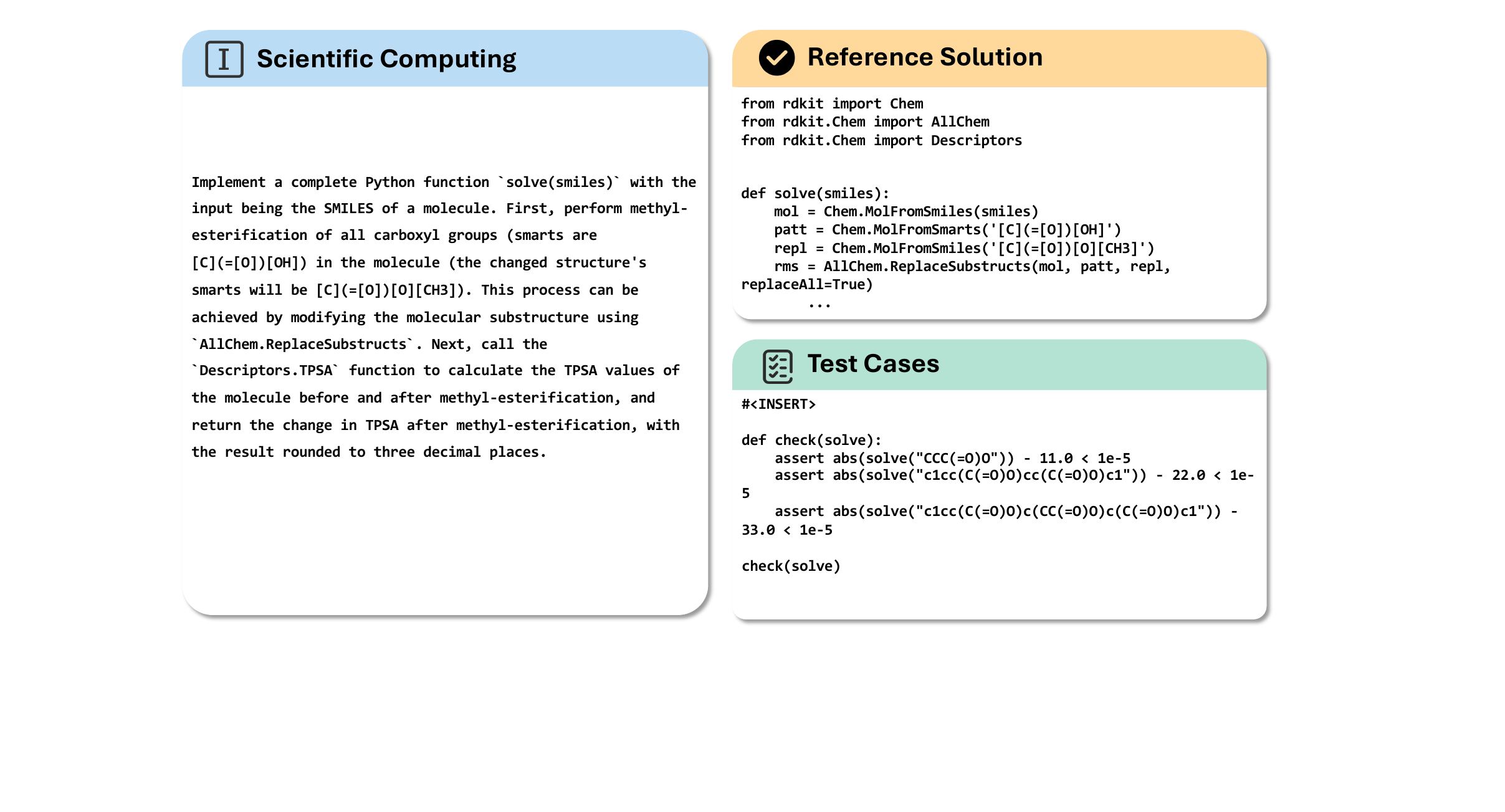} 
                 \includegraphics[width=0.40\linewidth]{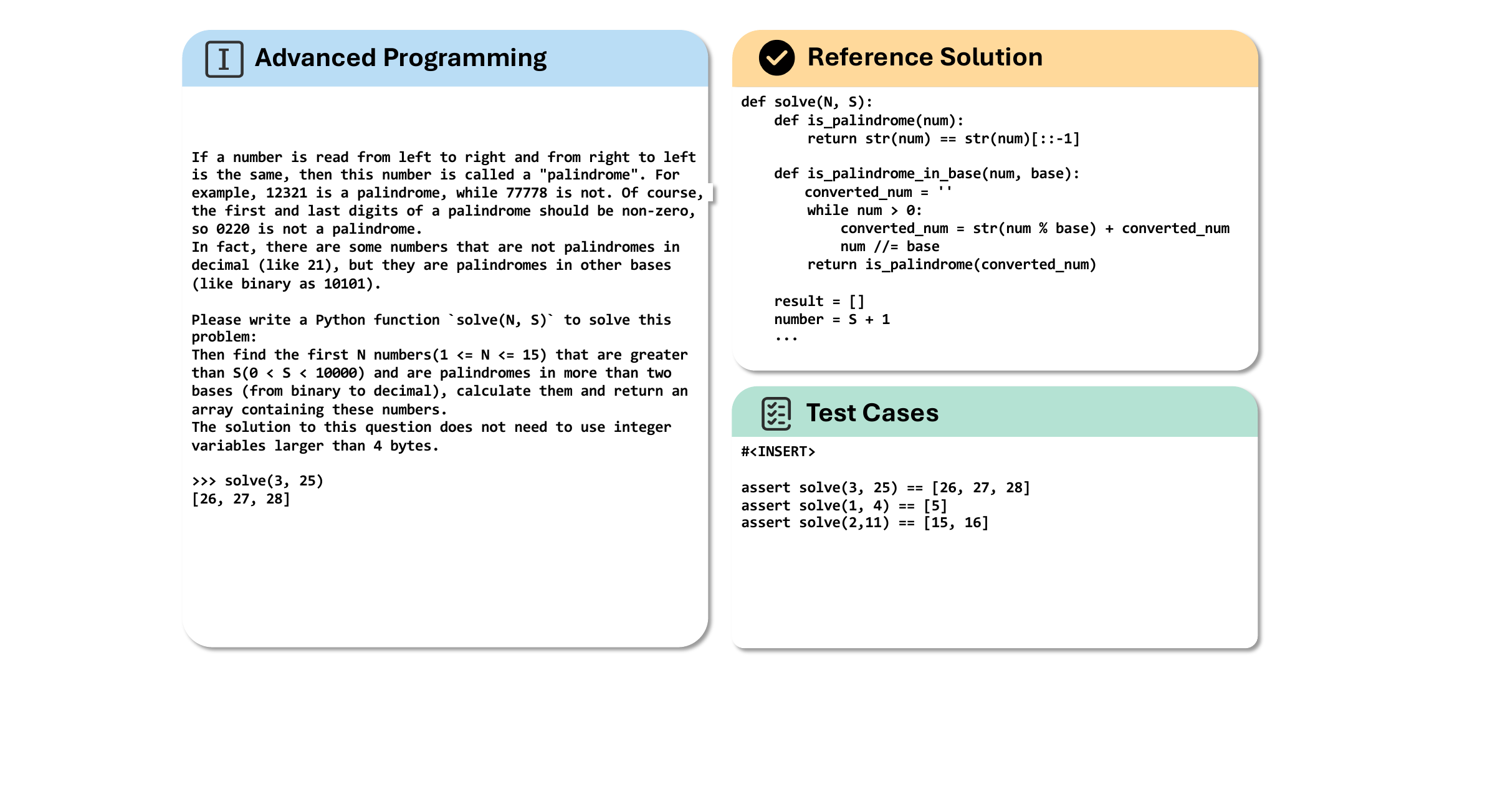}
                             \includegraphics[width=0.40\linewidth]{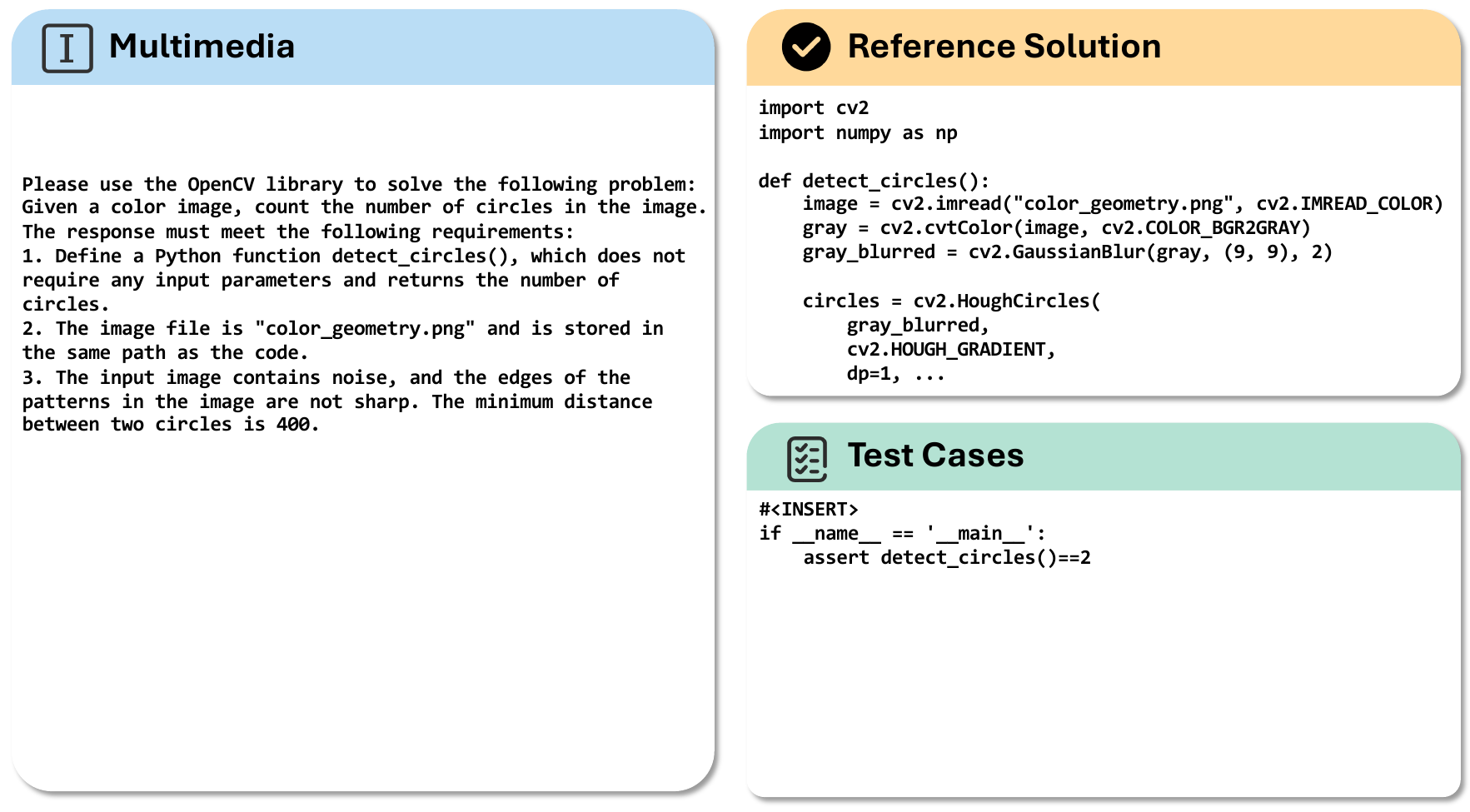}  
                             \includegraphics[width=0.40\linewidth]{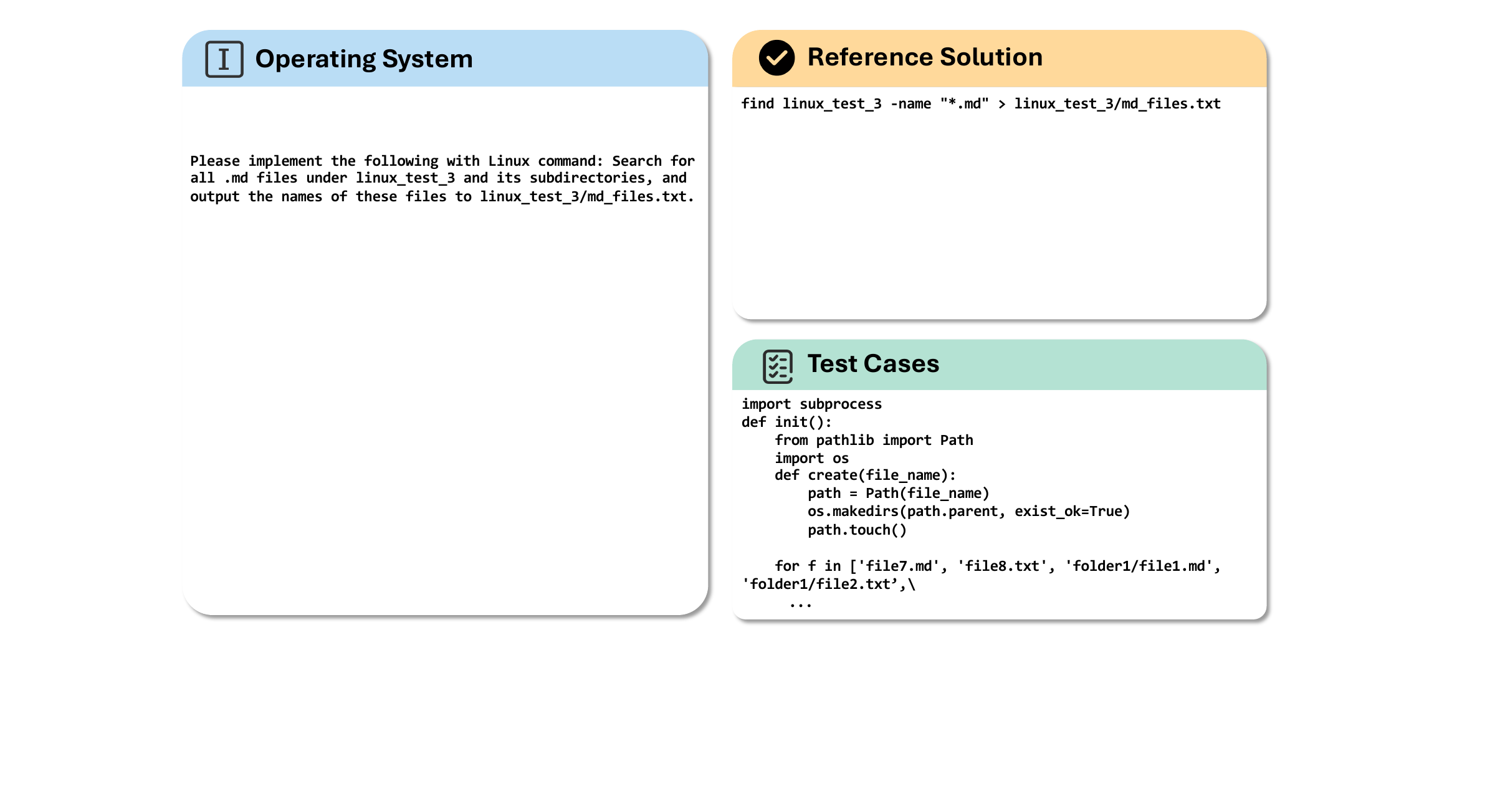}
                             \includegraphics[width=0.40\linewidth]{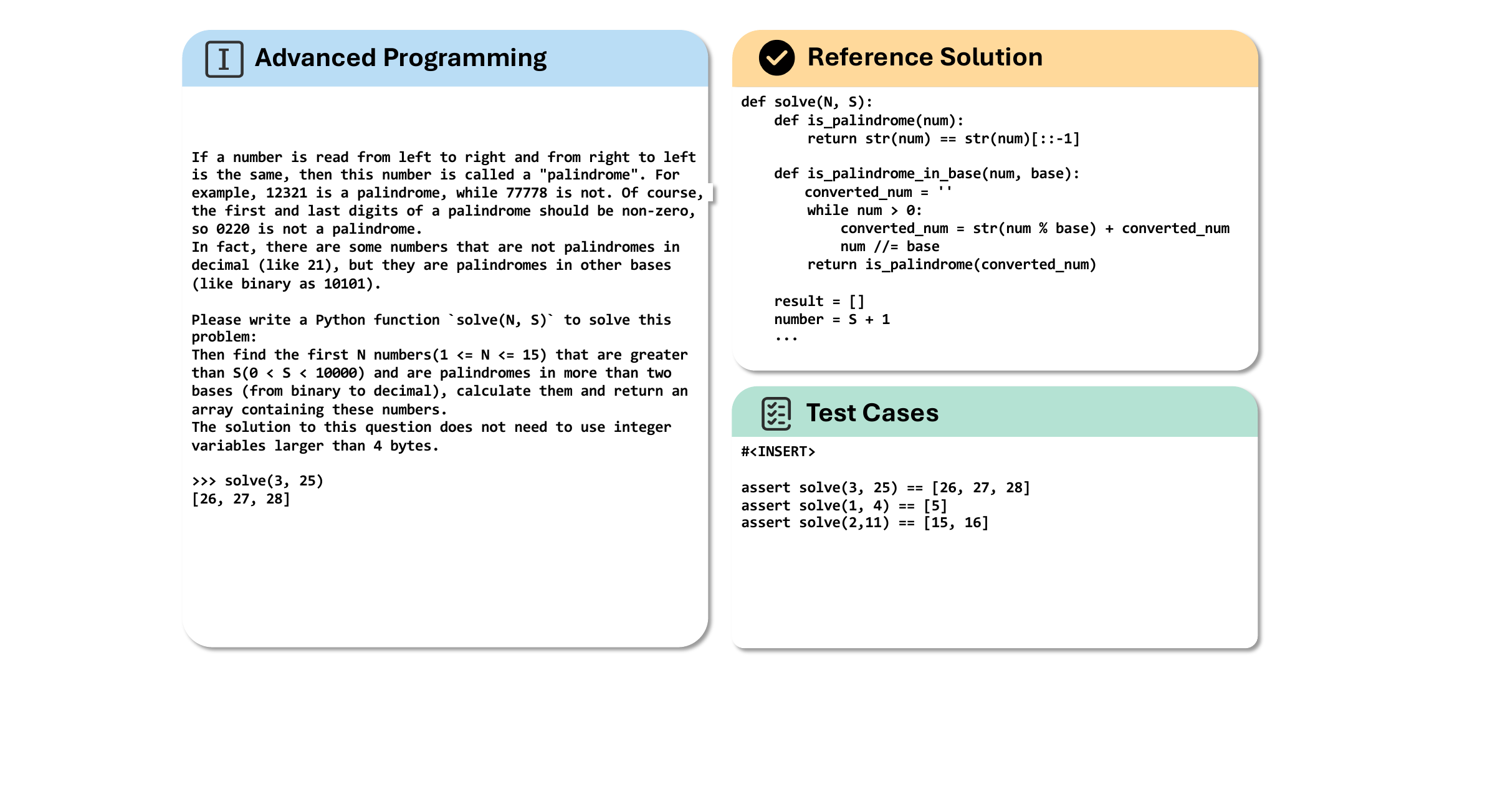}


    \caption{Visualization on some cases of our \benchmark{}.}
    \label{fig:level-python-12}
\end{figure}

\begin{table}

 \centering
\begin{tabular}{l|l}
 \toprule
 \textbf{Series Name} & \textbf{Model Name (Size)} \\
 \midrule
 \multirow{3}{*}{Llama} &  {CodeLlama-7B-Instruct (7B)} \\ 
 {} &  {CodeLlama-13B-Instruct (13B)} \\ 
 {} &  {CodeLlama-34B-Instruct (34B)} \\ 
 \midrule
 \multirow{4}{*}{Qwen2.5} &   {Qwen2.5-Coder-1.5B-Instruct (1.5B)} \\
 {} &  {Qwen2.5-Coder-7B-Instruct (7B)} \\
 {} &  {Qwen2.5-Coder-14B-Instruct (14B)} \\
 {} &  {Qwen2.5-Coder-32B-Instruct (32B)} \\ 
 \midrule
 \multirow{3}{*}{DeepSeek v1}&   {DeepSeek-Coder-1.3B-Instruct (1.3B)} \\
 {} &  {DeepSeek-Coder-6.7B-Instruct (6.7B)} \\
 {} &  {DeepSeekCoder-33B-Instruct (33B)} \\
 \midrule
 \multirow{3}{*}{DeepSeek v2} &  {DeepSeek-Coder-7B-Instruct-v1.5 (7B)} \\
 {} &  {DeepSeekCoder-v2-Lite-Instruct (16B)} \\
 {} &  {DeepSeekCoder-v2-Instruct (236B)} \\
 \midrule
 \multirow{2}{*}{OpenCoder} &  {OpenCoder-1.5B-Instruct (1.5B) } \\
 {} &  {OpenCoder-8B-Instruct (8B)} \\
 \bottomrule
 \end{tabular}
 \caption{Model Series discussed in Section~\ref{sec:scaling}.}
 \label{tab:series}
\end{table}
\begin{table}[htbp]
 \small
 \centering
 \resizebox{0.98\textwidth}{!}{
  \begin{tabular}{l|l}
 \toprule
 \textbf{Open-Sourced Model} & \textbf{Model Link} \\
 \midrule
CodeQwen1.5-7B-Instruct & \url{https://hf.co/Qwen/CodeQwen1.5-7B-Instruct} \\
 Qwen2.5-Coder-1.5B-Instruct & \url{https://hf.co/Qwen/Qwen2.5-Coder-1.5B-Instruct} \\
 Qwen2.5-Coder-7B-Instruct & \url{https://hf.co/Qwen/Qwen2.5-Coder-7B-Instruct} \\
 Qwen2.5-Coder-14B-Instruct & \url{https://hf.co/Qwen/Qwen2.5-Coder-14B-Instruct} \\
 Qwen2.5-Coder-32B-Instruct & \url{https://hf.co/Qwen/Qwen2.5-Coder-32B-Instruct} \\
 Qwen2.5-72B-Instruct & \url{https://hf.co/Qwen/Qwen2.5-72B-Instruct} \\
 \midrule
 CodeLlama-7B-Instruct & \url{https://hf.co/meta-llama/CodeLlama-7b-Instruct-hf} \\
 CodeLlama-34B-Instruct & \url{https://hf.co/meta-llama/CodeLlama-34b-Instruct-hf} \\
CodeLlama-13B-Instruct & \url{https://hf.co/meta-llama/CodeLlama-13b-Instruct-hf} \\
Llama3.1-70B-Instruct & \url{https://hf.co/meta-llama/Llama-3.1-70B-Instruct} \\
 \midrule
 DeepSeek-Coder-1.3B-Instruct & \url{https://hf.co/deepseek-ai/deepseek-coder-1.3b-instruct} \\
 DeepSeek-Coder-6.7B-Instruct & \url{https://hf.co/deepseek-ai/deepseek-coder-6.7b-instruct} \\
 DeepSeek-Coder-33B-Instruct & \url{https://hf.co/deepseek-ai/deepseek-coder-33b-instruct} \\
 DeepSeek-Coder-7B-Instruct-v1.5 & \url{https://hf.co/deepseek-ai/deepseek-coder-7b-instruct-v1.5} \\
 DeepSeek-Coder-V2-Lite-Instruct & \url{https://hf.co/deepseek-ai/DeepSeek-Coder-V2-Lite-instruct} \\
 DeepSeek-Coder-V2-Instruct & \url{https://hf.co/deepseek-ai/DeepSeek-Coder-V2-Instruct} \\
 \midrule
 OpenCoder-1.5B-Instruct & \url{https://hf.co/infly/OpenCoder-1.5B-Instruct} \\
 OpenCoder-8B-Instruct & \url{https://hf.co/infly/OpenCoder-8B-Instruct} \\
 Yi-Coder-9B-Chat & \url{https://hf.co/01-ai/Yi-Coder-9B-Chat} \\
 StarCoder2-15B-Instruct-v0.1 & \url{https://hf.co/bigcode/starcoder2-15b-instruct-v0.1} \\
 \bottomrule
 \end{tabular}
 }
 \caption{Open-sourced models adopted in our experiments.}
 \label{tab:open_source_model}
\end{table}

\begin{table}[htbp]
 \small
 \centering
 \resizebox{0.98\textwidth}{!}{
  \begin{tabular}{l|l}
 \toprule
 \textbf{Close-Sourced Model} & \textbf{API Entry} \\
 \midrule
 Claude-3.5-Sonnet & \url{https://www.anthropic.com/news/claude-3-5-sonnet} \\
 OpenAI o1-preview & \url{https://platform.openai.com/docs/models\#o1} \\
 OpenAI o1-mini & \url{https://platform.openai.com/docs/models\#o1} \\
 GPT 4o-0806 & \url{https://platform.openai.com/docs/models\#gpt-4o} \\
 DeepSeek-v2.5 & \url{https://www.deepseek.com} \\
 GLM-4-Plus & \url{https://open.bigmodel.cn/dev/api/normal-model/glm-4} \\
 Qwen-Max & \url{https://www.aliyun.com/product/bailian}\\
 \bottomrule
 \end{tabular}
 }
 \caption{Close-sourced models (APIs) adopted in our experiments.}
 \label{tab:api_model}
\end{table}
\subsection{Details of SandboxFusion}
\label{app:sandbox}
\subsubsection{Dataset Module}

The sandbox aims to provide a unified framework for most execution-based datasets, making it easier to add existing datasets or create new datasets based on them.
In particular, SandboxFusion implements many open-source code evaluation datasets, including HumanEval~\citep{chen2021evaluatinglargelanguagemodels}, MultiPL-E~\citep{multiple}, Shadow Humaneval~\citep{wei2023skyworkopenbilingualfoundation}, CodeContests~\citep{li2022competition}, MBPP~\citep{mbpp}, MBXP~\citep{mceval}, MHPP~\citep{dai2024mhppexploringcapabilitieslimitations}, CRUXEval~\citep{gu2024cruxeval}, NaturalCodeBench~\citep{zhang2024naturalcodebenchexaminingcodingperformance}, PAL-Math~\citep{gao2023pal}, verilog-eval~\citep{liu2023verilogevalevaluatinglargelanguage} and miniF2F~\citep{zheng2022minif2fcrosssystembenchmarkformal}. Besides,
as the judgment logic of most datasets is very similar,
to maximize code reuse,  we created two representative types of datasets: AutoEvalDataset and CommonOJDataset as follows:
\begin{itemize}

\item AutoEvalDataset. This type of dataset is primarily prepared for instruction-tuning models and can also be used to test the performance of pre-trained models through few-shot learning. The basic process is to extract the entire code block from the model's output, then concatenate the model code with pre-written test scripts and execute them, and finally check the return value of the executed program to determine whether the problem is passed. Note that we support \benchmark{} in this dataset mode.

\item CommonOJDataset. Most algorithmic competition problems belong to this category. These problems are language-agnostic and can be tested in any programming language. Most algorithmic competition datasets only store the problem description of the algorithmic competition problem and the test cases in the form of standard input and output. In the prompt generation stage, the problem description is combined with the instruction of the corresponding language as the complete prompt. The entire code block is extracted from the model's output and directly executed, and the standard output under the given standard input is compared to see if it matches the expectation.
\end{itemize}


\subsubsection{Sandbox Execution Module}

SandboxFusion covers a wide range of programming languages that have received attention in the field of code generation and provides a unified interface for executing them. Currently, SandboxFusion supports a total of 23 programming languages, including Python, C++, C\#, Go, Java, NodeJS, TypeScript, PHP, Rust, Bash, Lua, R, Perl, D, Ruby, Scala, Julia, Kotlin, Verilog, Lean, Swift, Racket, and CUDA. Together with these language compilers and interpreters, SandboxFusion also integrates many commonly used packages like PyTorch and TensorFlow for machine learning and a browser for front-end related application domains.
The main features of the Sandbox Execution Module are as follows:

\begin{itemize}

\item Our execution interface accepts a code snippet, language, and other input contexts, and returns the program's return code, standard output, and other information. This interface hides many language-specific implementation details from researchers and greatly improves the efficiency of data cleaning, compiler feedback, etc.

\item When dealing with incompatible versions of languages and packages, our principle is to select more advanced and popular versions. For example, while there are still many existing Lean codes written in Lean 3 on Github, we still use Lean 4 as the Lean community has fully transitioned to it. In cases where an older version of some package is indeed necessary, SandboxFusion also provides a function to run code in an isolated execution system to avoid affecting the main environment.

\item SandboxFusion provides functionality in some task-specific areas. For example, sandbox supports running CUDA and Python code in a GPU environment, making it possible to explore enhancing the performance of CUDA code. Another example is the supported Jupyter mode, which executes multiple code blocks in sequence, providing training signals for online user interaction.

\item 
SandboxFusion also has a built-in resource isolation functionality. Given limiting CPU and memory usage, it is possible to test a program's performance in a controlled environment, providing a reliable metric for program optimization. Its file system isolation allows testing scenarios that require file operations without affecting the host system, while network isolation enables multiple programs to bind to the same port simultaneously, preventing conflicts during concurrent testing. 
\end{itemize}

\subsection{Comparison with Other Sandboxes}
\label{app:compare}
\begin{table}[t]

\centering
\begin{tabularx}{\textwidth}{l|>{\raggedright\arraybackslash}X|>{\raggedright\arraybackslash}X|>{\raggedright\arraybackslash}X|>{\raggedright\arraybackslash}X}

\toprule
 & SandboxFusion & DifySandbox & MultiPL-E & MPLSandbox \\
\midrule
\# Datasets & 10+ & 0 & 2 & 0 \\
\midrule
\# Languages & 23 & 2 & 18 & 8 \\
\midrule
Security & namespace \& cgroup & seccomp & none & none(docker) \\
\midrule
API Type & HTTP \& SDK & HTTP & CLI & SDK \& CLI \\
\midrule
Deployment & Single Server & Single Server & CLI Tool & 1 Server + 8 Language Workers \\
\midrule 
Others & Jupyter Mode & Fine-Grained Security Limitation & - & Language Analysis \\
\bottomrule

\end{tabularx}
\caption{Comparison of different sandboxes.}

\label{tab:sandbox}

\end{table}

For comparison, we selected three representative sandboxes (i.e., DifySandbox, MultiPL-E, and MPLSandbox) as follows:
\begin{itemize}

\item DifySandbox~\citep{difysandbox} represents online code execution sandboxes with powerful and fine-grained security controls. 
In contrast, SandboxFusion, primarily used for internal model code evaluation, has lower security isolation requirements and thus only implements comprehensive isolation through basic Linux kernel capabilities like namespaces and cgroups, without providing fine-grained permission controls. Additionally, DifySandbox only supports Python and NodeJS runtime environments and has not implemented any code evaluation datasets. Compared to DifySandbox, SandboxFusion is optimized for large model evaluation and training scenarios, offering higher throughput and support for more datasets and languages.

\item MultiPL-E~\citep{cassano2023multiple} represents dataset-bundled code sandboxes. MultiPL-E built a problem translation pipeline and includes multiple language environments within its sandbox. However, it only serves the MultiPL-E dataset, and runs offline evaluation through CLI, making it impossible to integrate with online training tasks. Therefore, compared to such dataset-bundled execution environment sandboxes, SandboxFusion's advantages lie in its online interface and unified multi-dataset execution environment.

\item MPLSandbox~\citep{dou2024multiprogramminglanguagesandboxllms} represents research-oriented code sandboxes with rich code feedback signals (e.g., code execution, static analysis, and quality assessment). However, compared to SandboxFusion, MPLSandbox's deployment and usage are more complex. SandboxFusion requires only one docker image to run, while MPLSandbox needs to launch a service to execute each language. Furthermore, SandboxFusion provides enhanced security features, whereas MPLSandbox can only rely on container runtime environments for isolation, unable to prevent code interference between requests within containers.
\end{itemize}

More importantly, SandboxFusion can be used to evaluate the capabilities of LLM in various real-world code development scenarios. For instance, it supports testing desktop programs written with PyQt, web applications built with Flask, and neural networks with PyTorch or Tensorflow in  \benchmark{}. These codes are very common in real-world coding scenarios, but other sandboxes cannot support such testing. 
\end{document}